\documentclass{article}

\usepackage{arxiv}

\usepackage[utf8]{inputenc}
\usepackage[T1]{fontenc}
\usepackage{cite}
\usepackage{amsmath,amssymb,amsfonts}
\usepackage{mathptmx} 
\usepackage{graphicx}
\usepackage{textcomp}
\usepackage{booktabs}
\usepackage{multirow}
\usepackage{subcaption}
\usepackage{framed}
\usepackage{float}
\usepackage{placeins}
\usepackage[colorlinks=true,allcolors=blue]{hyperref}
\usepackage{algorithm}
\usepackage{algorithmic}

\usepackage{bm}
\usepackage{xspace}

\newcommand{\reqclar}{\texttt{request\_}\linebreak[1]\texttt{clarification}\xspace}

\def\BibTeX{{\rm B\kern-.05em{\sc i\kern-.025em b}\kern-.08em
    T\kern-.1667em\lower.7ex\hbox{E}\kern-.125emX}}

\title{Uncertainty Decomposition for Clarification Seeking in LLM Agents}

\author{%
  Gregory Matsnev \\
  AI Talent Hub, ITMO University\\
  Saint Petersburg 197101, Russia\\
  \texttt{gregory.matsnev@niuitmo.ru} \\
}
\date{\today}

\renewcommand{\shorttitle}{Uncertainty Decomposition for Clarification Seeking in LLM Agents}

\hypersetup{
  pdftitle={Uncertainty Decomposition for Clarification Seeking in LLM Agents},
  pdfauthor={Gregory Matsnev},
  pdfkeywords={clarification seeking, interactive benchmarks, LLM agents, prompt-based methods, uncertainty quantification},
}

\begin{document}
\maketitle

\begin{abstract}
Recent position papers argue that the classical aleatoric/epistemic uncertainty framework is insufficient for interactive large language model (LLM) agents and call for underspecification-aware, decomposed, and communicable uncertainty representations that can unlock new agent capabilities such as proactive clarification seeking and shared mental-model building.
Practical deployment constraints---black-box APIs, interactive latency budgets, and the absence of labeled trajectories---rule out logprob-based, multi-sampling, and training-based methods, leaving prompt-based estimation as the most viable family for surfacing such signals at deployment time.
We answer this call with a simple prompt-based decomposition that separates action confidence from request uncertainty ($u$), enabling the agent to ask for clarification when the task specification is ambiguous.
To evaluate it, we introduce two clarification-augmented benchmarks (WebShop-Clarification and ALFWorld-Clarification) in which 50\% of tasks are deliberately underspecified, and systematically compare the proposed decomposition against ReAct+UE and Uncertainty-Aware Memory (UAM) across five LLM backbones (GPT-5.1, DeepSeek-v3.2-exp, GLM-4.7, Qwen3.5-35B, GPT-OSS-120B) on these variants together with the standard WebShop, ALFWorld, and REAL benchmarks for fault detection.
Averaged across the five backbones, the proposed decomposition improves clarification F1 on ALFWorld-Clarification by 73\% over ReAct+UE and by 36\% over UAM, and leads clarification F1 on every backbone on WebShop-Clarification and on four of five backbones on ALFWorld-Clarification, indicating that the gains generalize beyond a single LLM.
\end{abstract}


\vspace{0.5em}
\noindent\textbf{Source code:} \url{https://github.com/PE51K/udcs-in-llm-agents}

\section{Introduction}
\label{sec:introduction}

Large language models (LLMs) trained on web-scale corpora have become general-purpose reasoners, exhibiting in-context learning~\cite{b1}, instruction following~\cite{b2}, and chain-of-thought reasoning~\cite{b3}, with capabilities that scale predictably with model size~\cite{b4}.
Building on these foundations, a growing body of work repurposes LLMs as the controllers of \emph{interactive agents} that plan, invoke tools, and act in external environments such as web interfaces and household simulations~\cite{b5},~\cite{b6},~\cite{b7},~\cite{b8}.
Unlike single-turn question answering, these agents operate under partial observability: they receive underspecified natural-language instructions, observe noisy environment states, and must chain multiple reasoning steps to complete a task.
Small errors at early steps -- misinterpreting an ambiguous request, over-trusting a noisy observation, or selecting a suboptimal action -- can propagate along the trajectory and produce a confidently wrong outcome~\cite{b9},~\cite{b10}.

Uncertainty estimation is a natural tool for mitigating such failures~\cite{b11},~\cite{b12}.
However, recent position papers argue that existing uncertainty frameworks are fundamentally inadequate for interactive agents.
Kirchhof et al.~\cite{b13} demonstrate that the traditional aleatoric/epistemic dichotomy breaks down in agent settings: when a chatbot decides whether to ask a follow-up question, the same uncertainty can be classified as aleatoric (irreducible at the current time point) or epistemic (reducible by asking), depending on the modeler's perspective.
They propose three research directions: underspecification uncertainties that arise when users do not provide complete information, interactive learning through follow-up questions to reduce context uncertainty, and rich output uncertainties communicated as natural language rather than scalar scores.
Kim et al.~\cite{b14} complement this vision with agentic interpretability -- a paradigm where agents proactively assist human understanding through multi-turn interaction, developing and leveraging mutual mental models.
Together, these position papers call for uncertainty methods that are decomposed by source, communicable to users, and capable of enabling new agent capabilities beyond simple abstention.

Practical deployment constraints further shape the design space.
Black-box API access precludes logprob-based methods~\cite{b16},~\cite{b17}.
Multi-sampling at every agent step introduces prohibitive latency and cost in long-horizon settings~\cite{b18},~\cite{b17}.
Training-based calibrators require labeled data and model access~\cite{b19},~\cite{b20}, and white-box methods that read internal hidden states~\cite{b15} are similarly precluded by closed APIs.
This leaves prompt-based methods -- where the agent emits uncertainty estimates as structured text alongside its actions -- as the most practically viable approach for real-world agentic deployments, despite their known limitations.

Existing prompt-based methods, however, produce a single scalar confidence per step.
This conflates fundamentally different sources of uncertainty.
An agent may report low confidence because the action is difficult (e.g., many similar products to choose from) or because the user request is ambiguous (e.g., ``find me a shirt'' without specifying color or size).
These two situations call for different responses: the former suggests the agent should proceed cautiously, while the latter suggests it should ask the user for clarification.

In this paper, we propose a simple decomposition of prompt-based uncertainty into two components:
\begin{itemize}
    \item \textbf{Action confidence} ($c_t$): the agent's confidence that its chosen action moves toward task completion, given the current understanding of the task.
    \item \textbf{Request uncertainty} ($u_t$): the agent's estimate of whether the user's goal is fully specified, ranging from 0 (fully specified) to 1 (critical details missing).
\end{itemize}
Unlike clarifiers that require training on labeled trajectories, our method is prompt-only and runs on black-box LLMs; we characterize both the capabilities and the limitations of eliciting this decomposition through prompting alone.

We evaluate this decomposition against two prompt-based baselines -- ReAct with a simple uncertainty estimation prompt suffix (ReAct+UE) and Uncertainty-Aware Memory (UAM)~\cite{b21} -- across five LLM backbones (GPT-5.1, DeepSeek-v3.2-exp, GLM-4.7, Qwen3.5-35B, GPT-OSS-120B), on three standard interactive benchmarks (WebShop, ALFWorld, REAL) and two clarification-augmented variants (WebShop-Clarification, ALFWorld-Clarification) in which 50\% of tasks are deliberately underspecified, so the agent must recognize the gap and ask the user.
We also run a sensitivity analysis over the $u_t$ clarification threshold.

Our contributions are:
\begin{enumerate}
    \item A comparative analysis of uncertainty estimation approaches for LLM agents, showing that practical deployment constraints leave prompt-based methods as the only viable family and motivating a systematic study of them for proactive agent capabilities.
    \item Two clarification-augmented benchmarks (WebShop-Clarification and ALFWorld-Clarification) in which 50\% of tasks are deliberately underspecified, enabling evaluation of clarification seeking as a binary classification task.
    \item A prompt-based decomposition method that separates action confidence from request uncertainty, enabling proactive clarification seeking, evaluated against ReAct+UE and UAM across five LLM backbones on the clarification-augmented benchmarks.
    \item An empirical evaluation of prompt-based methods on fault detection across standard interactive benchmarks, surfacing their capabilities and limitations and promising directions for future work.
\end{enumerate}

\section{Related Work}
\label{sec:related}

We review prior work in three threads.
We first cover single-turn uncertainty estimation methods for LLMs, where most of the technical machinery originates.
We then turn to methods that propagate per-step uncertainty along multi-step agent trajectories, and summarize the families in a comparison table that motivates our focus on prompt-based approaches.
Finally, we discuss uncertainty decomposition and clarification seeking, which form the immediate context for the proposed method.

\subsection{Uncertainty Estimation in LLMs}
\label{sec:rw-uq-llm}

Uncertainty estimation methods for LLMs span several families~\cite{b12},~\cite{b17}.
\emph{Multi-sampling} methods draw multiple responses and measure their disagreement: self-consistency~\cite{b18} selects the majority answer among chain-of-thought samples, while semantic entropy~\cite{b43},~\cite{b22}, kernel language entropy~\cite{b23}, semantic-embedding variants~\cite{b40}, and similarity-based dispersion measures for black-box NLG~\cite{b44} quantify disagreement in semantic space, with joint entropy modeling over LLM and tool contributions extending these measures to tool-using QA systems~\cite{b25}, and input clarification ensembling~\cite{b26} decomposing uncertainty by generating and ensembling multiple clarified versions of the input.
\emph{Logprob-based} methods exploit the model's output token probabilities: perplexity~\cite{b46} and predictive entropy~\cite{b45} aggregate per-token log-probabilities over a generation, while relevance-weighted variants such as Shifting Attention to Relevance~\cite{b47} reweight tokens by semantic importance before aggregating.
\emph{Prompt-based} methods ask the model to express confidence directly~\cite{b24}, optionally calibrating the elicited scores through multi-agent deliberation~\cite{b41}.
\emph{Training-based} methods learn a separate model -- a probe or recalibrator over the base model's output logprobs and/or internal hidden representations, supervised by answer-correctness labels: ProbeCal~\cite{b19} recalibrates a tool-using agent's internal token probabilities, while MICE~\cite{b20} trains a classifier on model-internal activations; some such methods further leverage internal hidden states from intermediate layers~\cite{b15}.
Engineering toolkits such as LM-Polygraph~\cite{b16} provide unified implementations across these families.

These methods primarily target single-turn prediction.
For multi-step agents, uncertainty arises and evolves at each think-act-observe step, requiring propagation mechanisms.

\subsection{Uncertainty Propagation in Agent Trajectories}
\label{sec:rw-prop}

Two recent frameworks address step-wise uncertainty propagation.
SAUP~\cite{b9} attaches per-step uncertainty estimates using plug-in estimators and propagates them via HMM-based situational weights, achieving up to 20\% AUROC improvement over final-step-only baselines.
UProp~\cite{b10} formalizes propagation using pointwise mutual information over sampled trajectories.
Both methods, however, rely on resources that are typically unavailable in black-box API deployments---multi-sampling, output logprobs, or labeled trajectories for training---making them impractical in this setting.

BrowseConf~\cite{b27} uses confidence scores for test-time scaling in web agents, dynamically allocating compute based on the agent's self-assessed uncertainty.
Uncertainty-Aware Memory (UAM)~\cite{b21} includes the agent's confidence score and natural-language explanation in the action history, allowing subsequent steps to reason about accumulated uncertainty.
This prompt-based propagation requires no additional API calls, making it suitable for practical deployment.

\subsection{Comparative Overview of Approaches}
\label{sec:rw-overview}

Table~\ref{tab:comparison} provides a systematic comparison of uncertainty estimation approaches for LLM agents.
Methods differ in their requirements (logprob access, model-internals access, multiple inference passes, training data) and capabilities (multi-step support, clarification).
The full taxonomy is included for completeness and to motivate the design choices we formalize in Section~\ref{sec:problem}.

\subsection{Uncertainty Decomposition and Clarification}
\label{sec:rw-decomp}

Position papers argue that the aleatoric/epistemic split is insufficient for interactive agents~\cite{b13},~\cite{b28}, and empirical analyses of uncertainty sources in LLMs and multimodal models reach similar conclusions~\cite{b42}.
Kirchhof et al.~\cite{b13} introduce \emph{underspecification uncertainty} -- uncertainty arising when users do not provide complete information -- as a category distinct from model knowledge gaps, and advocate for interactive learning via follow-up questions and rich natural-language uncertainty outputs.
Smith et al.~\cite{b28} formalize that the aleatoric/epistemic boundary is inherently modeler-dependent.
These arguments have begun to be operationalized in concrete agent systems.
SAGE-Agent~\cite{b29} trains a POMDP-guided clarifier with GRPO, and Hao et al.~\cite{b31} trigger human-in-the-loop refinement in GUI agents when uncertainty is high; in parallel, decision-theoretic frameworks such as DeLLMa~\cite{b32} and PlanU~\cite{b33} connect uncertainty to action selection through utility modeling.
A recurring caveat across this line of work is that agents tend to be systematically overconfident~\cite{b30}, limiting the reliability of the very signals these methods depend on.

\begin{table*}
\centering
\caption{Comparative Overview of Uncertainty Estimation Approaches for LLM Agents.
Requirements indicate what each method needs beyond a single forward pass.
``Multi-step'' indicates native support for trajectory-level uncertainty.
``Clarification'' indicates whether the method can trigger user clarification.
Prompt-based methods are the only family requiring neither logprob access nor model internals nor multiple inference passes, motivating the experimental focus of this paper.}
\label{tab:comparison}
\setlength{\tabcolsep}{3pt}
\begin{tabular}{|p{60pt}|p{100pt}|p{85pt}|p{40pt}|p{50pt}|p{90pt}|}
\hline
\textbf{Family} & \textbf{Representative Methods} & \textbf{Requirements} & \textbf{Multi-step} & \textbf{Clarification} & \textbf{Key Limitation} \\
\hline
Logprob-based & Perplexity~\cite{b46}, Pred.\ Entropy~\cite{b45}, SAR~\cite{b47}, LM-Polygraph~\cite{b16} & Output token logprobs & No & No & Not applicable to black-box APIs \\
\hline
Multi-sampling & Self-Consistency~\cite{b18}, Sem.\ Entropy~\cite{b22}, KLE~\cite{b23}, Sem.\ Embed.~\cite{b40}, Deliberation~\cite{b41}, Clarif.\ Ensembling~\cite{b26} & $N$ forward passes per step & Partial$^{*}$ & No & $N\times$ cost/latency; not applicable to black-box APIs \\
\hline
Training-based & ProbeCal~\cite{b19}, MICE~\cite{b20}, Int.\ Belief~\cite{b15}, SAGE-Agent~\cite{b29}, GUI-Agent~\cite{b34} & Logprobs or model internals, labeled data & Yes & Yes$^{\dagger}$ & Requires labeled data and training \\
\hline
Prompt-based & ReAct+UE~\cite{b24}, UAM~\cite{b21}, BrowseConf~\cite{b27}, Proposed & Single forward pass, prompt only & Yes & Yes$^{\ddagger}$ & Overconfidence, capability dilution \\
\hline
\multicolumn{6}{p{250pt}}{$^{*}$ SAUP~\cite{b9} and UProp~\cite{b10} add multi-step propagation atop multi-sampling. $^{\dagger}$ SAGE-Agent only. $^{\ddagger}$ Proposed method only.}\\
\end{tabular}
\end{table*}

\section{Problem Statement}
\label{sec:problem}

We formalize the problem of uncertainty estimation for interactive LLM agents under practical deployment constraints.
We first define the agent setting and notation, then argue that these constraints leave prompt-based methods as the only viable family, scoping our experimental comparison accordingly.
We then introduce two evaluation objectives together with the metrics used to measure them: fault detection, the standard task that uncertainty methods are designed to solve, and clarification seeking, the proactive capability the proposed decomposition is intended to enable.
Finally, we describe the two existing prompt-based methods, ReAct+UE and UAM, against which the proposed method is compared; we introduce them here rather than in Section~\ref{sec:method} because they are prior work, not contributions of this paper.

\subsection{Task Setting and Notation}
\label{sec:ps-setting}

An LLM agent operates in an environment with observation space $\mathcal{O}$ and action space $\mathcal{A}$.
A task is specified by a natural-language goal $g \in \mathcal{G}$.
At each step $t = 1, \ldots, T$, the agent receives an observation $o_t \in \mathcal{O}$ and must produce an action $a_t \in \mathcal{A}$.
The agent is implemented as an LLM module $\pi$ that generates structured output:
\begin{equation}
(r_t, a_t, s_t) \sim \pi(\cdot \mid g, H_t, o_t; \phi),
\label{eq:llm-module}
\end{equation}
where $r_t$ is the chain-of-thought reasoning, $s_t \in [0,1]^k$ is a vector of uncertainty signals, $H_t = \{(o_i, r_i, a_i, s_i)\}_{i=1}^{t-1}$ is the interaction history retained in context, and $\phi$ denotes the instrumentation prompt that specifies which uncertainty signals the agent should emit.
A trajectory $\tau = (o_1, a_1, \ldots, o_T, a_T)$ receives a binary success label $y(\tau) \in \{0,1\}$ determined by the environment.
For tasks drawn from clarification-augmented benchmarks, each task also carries an underspecification label $z \in \{0, 1\}$, where $z=1$ indicates the goal is underspecified.

\subsection{Why Prompt-Based Methods?}
\label{sec:ps-why}

The families in Table~\ref{tab:comparison} rely on resources that are typically unavailable when deploying an agent on top of a commercial LLM API.
As Oh et al.~\cite{b17} observe, ``probability-based methods cannot be applied to most frontier LLMs'' and ``consistency-based methods become infeasible due to their prohibitively high inference cost in long-horizon, multi-turn settings.'' Logprob-based methods require access to output token probabilities, and training-based methods that leverage internal hidden states require white-box access -- neither of which most production APIs expose.
Multi-sampling methods scale compute and latency as $\mathcal{O}(N T)$ across a trajectory of length $T$, which is rarely acceptable at interactive-agent budgets.
Training-based methods require labeled trajectories to train a separate probe or recalibrator~\cite{b19},~\cite{b20}, and in some cases the ability to fine-tune the underlying model itself~\cite{b29}.
Prompt-based methods are the only family that runs in a single forward pass on a black-box API, and they natively compose with multi-step reasoning since the uncertainty signal is just part of the generated text.
We therefore restrict our experimental comparison to the prompt-based family (ReAct+UE, UAM, and the proposed method) and treat the other families as the context that motivates this restriction.

\subsection{Fault Detection Objective}
\label{sec:ps-faultdet}

Fault detection -- using a trajectory's uncertainty signals to predict whether it will fail -- is the conventional task on which uncertainty estimation methods are evaluated.
Given per-step uncertainty signals $\{s_t\}_{t=1}^{T}$, we obtain a trajectory-level score $S(\tau) \in [0,1]$ via an aggregation function $\mathrm{Agg}$, and evaluate how well $S$ predicts the binary success $y$.
Concretely, with $N$ trajectories and predictions $\hat{p}_i = S(\tau_i)$:
\begin{gather}
\text{ROC-AUC} = \Pr\!\left(\hat{p}_{i} > \hat{p}_{j} \mid y_i=1, y_j=0\right), \label{eq:roc} \\
\text{Brier} = \frac{1}{N} \sum_{i=1}^{N} (\hat{p}_i - y_i)^2, \label{eq:brier} \\
\text{ECE} = \sum_{b=1}^{B} \frac{|I_b|}{N}\,\bigl|\mathrm{acc}(I_b) - \mathrm{conf}(I_b)\bigr|, \label{eq:ece}
\end{gather}
where $\Pr(\cdot)$ denotes the probability over a uniformly drawn pair $(i, j)$ with $y_i = 1$ and $y_j = 0$, $I_b$ is the set of trajectories whose predicted score falls in confidence bin $b$, $\mathrm{acc}(I_b)$ is the empirical success rate in bin $b$, and $\mathrm{conf}(I_b)$ is the mean predicted score in bin $b$.
ROC-AUC captures how well $S$ discriminates failing from succeeding trajectories, ECE captures how closely its values match empirical success rates (calibration), and the Brier score reflects both discrimination and calibration quality jointly.

\subsection{Clarification-Seeking Objective}
\label{sec:ps-clarif}

Clarification seeking is the distinctive objective we evaluate beyond standard fault detection: rather than only scoring its own reliability, the agent must recognize an underspecified goal and act on it.
On clarification-augmented benchmarks the agent has the option to emit the special action $a_t =$\,\reqclar.
Let $D(\tau) = \mathbb{1}[\exists t:\, a_t =$\,\reqclar\unskip$]$ be the trajectory-level indicator that the agent asked for clarification.
We evaluate $D$ as a binary predictor of the underspecification label $z$.
Over the $N$ trajectories, define the index sets
\begin{align}
\mathrm{TP} &= |\{i : D(\tau_i) = 1,\ z_i = 1\}|, \notag \\
\mathrm{FP} &= |\{i : D(\tau_i) = 1,\ z_i = 0\}|, \notag \\
\mathrm{FN} &= |\{i : D(\tau_i) = 0,\ z_i = 1\}|, \notag \\
\mathrm{TN} &= |\{i : D(\tau_i) = 0,\ z_i = 0\}|. \notag
\end{align}
The classification metrics are then
\begin{gather}
\text{Precision} = \frac{\mathrm{TP}}{\mathrm{TP}+\mathrm{FP}}, \label{eq:precision} \\
\text{Recall} = \frac{\mathrm{TP}}{\mathrm{TP}+\mathrm{FN}}, \label{eq:recall} \\
\text{F1} = \frac{2\,\text{Precision}\cdot\text{Recall}}{\text{Precision}+\text{Recall}}, \label{eq:f1} \\
\text{Accuracy} = \frac{\mathrm{TP}+\mathrm{TN}}{N}. \label{eq:acc}
\end{gather}
Because $D$ is a hard decision derived from the agent's action sequence, Precision/Recall/F1/Accuracy depend on the method and threshold but not on the aggregation function $\mathrm{Agg}$.

\subsection{Baseline: ReAct + Uncertainty Estimation}
\label{sec:ps-react}

The first prompt-based baseline is ReAct+UE, formulated by Zhang et al.~\cite{b21}.
It augments the standard ReAct agent~\cite{b5} with per-step confidence elicitation in the spirit of verbalized uncertainty estimation~\cite{b24}.
At each step $t$, the agent outputs:
\begin{align}
&\texttt{<think>}\ r_t\ \texttt{</think>} \notag \\
&\texttt{<action>}\ a_t\ \texttt{</action>} \notag \\
&\texttt{<confidence>}\ c_t \in [0,1]\ \texttt{</confidence>} \label{eq:react} \\
&\texttt{<explanation>}\ e_t\ \texttt{</explanation>} \notag
\end{align}
Crucially, $c_t$ and $e_t$ are \emph{not} written back into the agent's history for subsequent steps.
The agent has no memory of its past uncertainty, preventing it from reasoning about accumulated confidence.
The confidence elicitation instructions are appended to every user turn and ask the agent to report a confidence value in $[0,1]$ together with a natural-language explanation of what makes it confident, what concerns it has, what information might be missing, and what alternative actions it considered.
The full prompt for ReAct+UE is given in Appendix~\ref{app:prompt-react}.

\subsection{Baseline: Uncertainty-Aware Memory}
\label{sec:ps-uam}

The second prompt-based baseline is Uncertainty-Aware Memory (UAM), introduced by Zhang et al.~\cite{b21}.
UAM uses the same output format and confidence elicitation prompt as ReAct+UE, but now propagates the confidence score and explanation through the agent's history:
\begin{equation}
H_t^{\text{UAM}} = \{(o_i, r_i, a_i, c_i, e_i)\}_{i=1}^{t-1}.
\label{eq:uam_history}
\end{equation}
This allows the agent to reason about past uncertainty levels, adjust confidence based on accumulated evidence, and detect patterns of decreasing or increasing certainty.
Following Zhang et al.~\cite{b21}, we use their ``Variant B: Semantic Propagation''.
UAM therefore differs from ReAct+UE only in whether uncertainty is retained in context; it shares the same single-scalar confidence signal $s_t = c_t$.
The full prompt for UAM is given in Appendix~\ref{app:prompt-uam}.

\section{Proposed Method}
\label{sec:method}

\begin{figure}[t]
\centering
\includegraphics[width=0.65\textwidth]{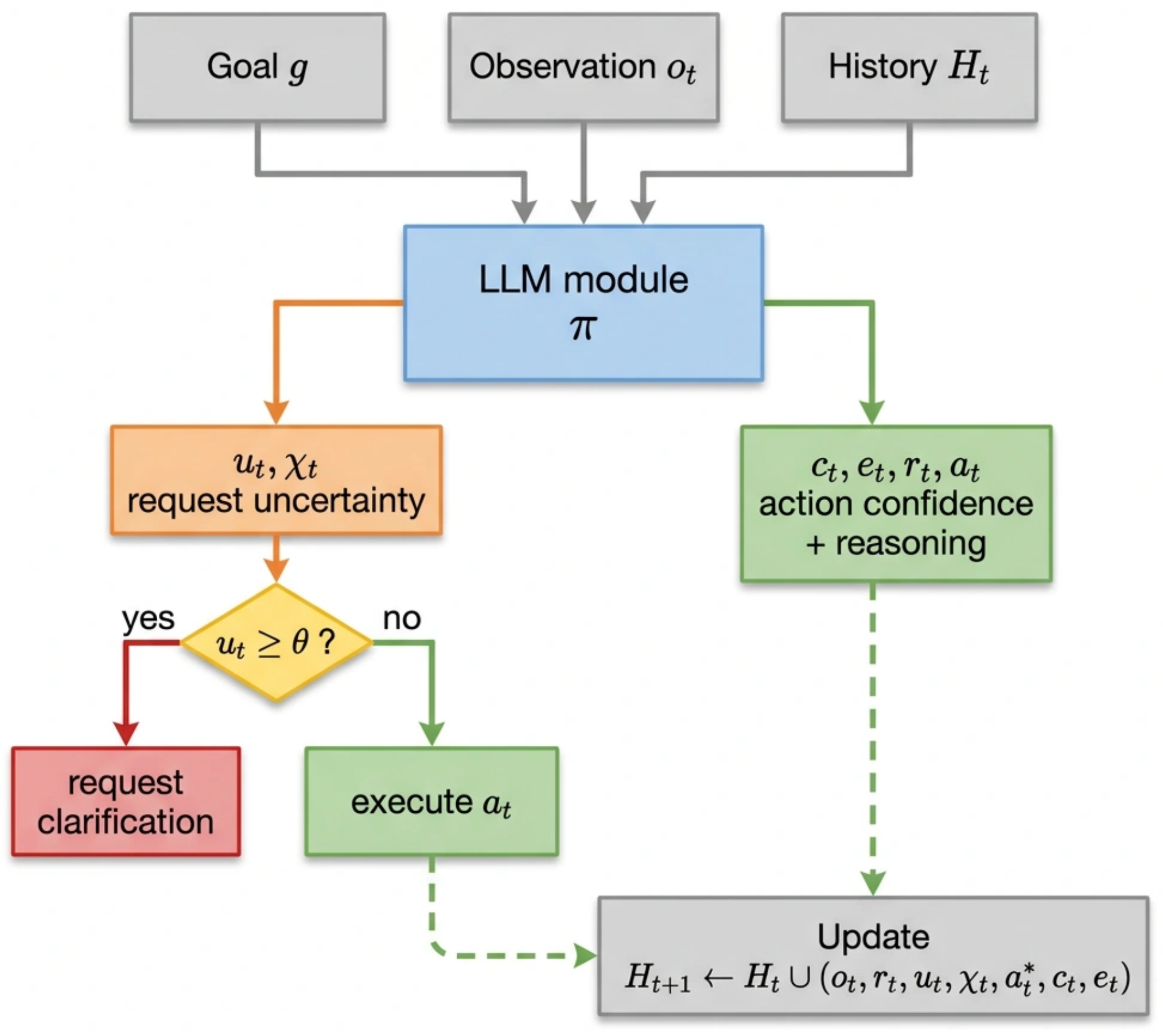}
\caption{Proposed method at step $t$.
The LLM module $\pi$ (blue) consumes the goal $g$, current observation $o_t$, and history $H_t$ in one forward pass and emits two uncertainty signals: request uncertainty $u_t$ with explanation $x_t$ (orange), and action confidence $c_t$ with explanation $e_t$ alongside the reasoning $r_t$ and proposed action $a_t$ (green).
The deterministic routing test $u_t \ge \theta$ switches between \reqclar and execution of $a_t$.
All emitted fields are appended to the history for subsequent steps.
Assessing $u_t$ before emitting $a_t$ ensures underspecification is caught prior to any action, giving the agent a dedicated channel for goal ambiguity that a single confidence score cannot provide.}
\label{fig:method}
\end{figure}

We now introduce the proposed method.
Its defining feature is a decomposition of the single confidence scalar used by ReAct+UE and UAM into two semantically distinct signals: an action confidence $c_t$ and a request uncertainty $u_t$.
Figure~\ref{fig:method} sketches the per-step data flow and Algorithm~\ref{alg:proposed} gives the step-level pseudocode for the proposed method.
The remainder of this section describes the two signals, the clarification trigger, the history propagation, and the trajectory-level aggregation strategies.
The full prompt for the proposed method is given in Appendix~\ref{app:prompt-proposed}.

\begin{algorithm}[t]
\caption{Proposed step: decomposed prompt-based uncertainty with clarification routing.
The LLM module $\pi$ consumes the goal $g$, current observation $o_t$, and history $H_t$ in a single forward pass, and emits reasoning $r_t$, request uncertainty $u_t$ with explanation $x_t$, action $a_t$, and action confidence $c_t$ with explanation $e_t$.
The $u_t \ge \theta$ test deterministically routes the agent to \reqclar when the goal is judged underspecified.
All emitted fields are appended to history so later steps can reason over past uncertainty.}
\label{alg:proposed}
\begin{algorithmic}[1]
\REQUIRE goal $g$, observation $o_t$, history $H_t$, threshold $\theta$
\ENSURE action $a_t^{\star}$, updated history $H_{t+1}$
\STATE $(r_t, u_t, x_t, a_t, c_t, e_t) \sim \pi(\cdot \mid g, H_t, o_t; \phi_{\text{proposed}})$
\IF{$u_t \ge \theta$}
  \STATE $a_t^{\star} \leftarrow$\,\reqclar
\ELSE
  \STATE $a_t^{\star} \leftarrow a_t$
\ENDIF
\STATE $H_{t+1} \leftarrow H_t \cup \{(o_t, r_t, u_t, x_t, a_t^{\star}, c_t, e_t)\}$
\RETURN $a_t^{\star}, H_{t+1}$
\end{algorithmic}
\end{algorithm}

\subsection{Decomposed Uncertainty Signals}

The proposed method extends UAM by replacing the single $c_t$ scalar with the two-scalar signal $(u_t, c_t)$.
At each step the agent emits, in order:
\begin{align}
&\texttt{\footnotesize <think>}\,r_t\,\texttt{\footnotesize </think>} \notag \\
&\texttt{\footnotesize <u\_request>}\,u_t \in [0,1]\,\texttt{\footnotesize </u\_request>} \notag \\
&\texttt{\footnotesize <u\_request\_explanation>}\,x_t\,\texttt{\footnotesize </\ldots>} \label{eq:decomp} \\
&\texttt{\footnotesize <action>}\,a_t\,\texttt{\footnotesize </action>} \notag \\
&\texttt{\footnotesize <confidence>}\,c_t \in [0,1]\,\texttt{\footnotesize </confidence>} \notag \\
&\texttt{\footnotesize <explanation>}\,e_t\,\texttt{\footnotesize </explanation>} \notag
\end{align}

The two signals serve distinct purposes and have different intellectual origins.

\paragraph{Action confidence $c_t$.} Estimates how likely the chosen action $a_t$ is to make progress toward task completion, conditioned on the agent's current understanding of the goal.
This signal is unchanged from the baselines; we retain it so that the proposed method can be directly compared against them on the fault detection task.

\paragraph{Request uncertainty $u_t$.} Estimates the degree to which the user's goal is underspecified, on a three-point anchored scale:
\begin{itemize}
    \item $u_t = 0$: The goal fully specifies every relevant parameter; there is exactly one correct interpretation.
    \item $u_t = 0.5$: At least one parameter is left open; the user likely has a specific preference that is not stated.
    \item $u_t = 1$: Critical details are missing; many equally valid interpretations exist.
\end{itemize}
This scale operationalizes the ``underspecification uncertainty'' category argued for by Kirchhof et al.~\cite{b13} and conceptually mirrors input clarification ensembling~\cite{b26}, but realizes it at the prompt level rather than through multi-sampling.

\paragraph{Clarification trigger.} The agent is instructed that when $u_t \ge \theta$ (with $\theta=0.5$ unless stated otherwise), the action it emits must be \reqclar.
This creates a direct, deterministic link between the uncertainty estimate and an observable behavior, and is what allows the decomposition to be evaluated as a binary classifier over the underspecification label $z$.

\paragraph{History propagation.} Both $u_t, x_t, c_t, e_t$ are included in the history, following UAM's semantic propagation:
\begin{equation}
H_t^{\text{proposed}} = \{(o_i, r_i, u_i, x_i, a_i, c_i, e_i)\}_{i=1}^{t-1}.
\label{eq:decomp_history}
\end{equation}
Compared to UAM's history in Eq.~(\ref{eq:uam_history}) this adds the $(u_i, x_i)$ pair so the agent can reason about request-level uncertainty across steps.

\paragraph{Output field ordering.} Within the structured output of Eq.~(\ref{eq:decomp}), the agent emits the $u_t$ assessment before the action $a_t$.
The agent first decides whether the goal is sufficiently specified, and this judgment is allowed to influence the action choice (clarification vs.\ task action).

\subsection{Prompt}

The proposed method instructs the agent, before selecting an action, to assess request uncertainty $u_t$ on the anchored 0/0.5/1 scale defined above and to explain its assessment in a free-text field.
The full prompt for the proposed method is given in Appendix~\ref{app:prompt-proposed}.

\subsection{Trajectory-Level Aggregation}
\label{sec:method-agg}

Every method except ReAct+UE already aggregates uncertainty across steps implicitly, by propagating the per-step estimates through the agent's history (Eqs.~(\ref{eq:uam_history}) and~(\ref{eq:decomp_history})).
For completeness and comparability with other work on LLM-agent uncertainty, we additionally apply, on top of this built-in propagation, a set of explicit trajectory-level aggregation functions $\mathrm{Agg}$ to the per-step signals.
For action confidence $c_1, \ldots, c_T$ we consider four strategies:
\begin{align}
S_c^{\text{last}} &= c_T, \label{eq:last} \\
S_c^{\text{avg}} &= \frac{1}{T}\sum_{t=1}^{T} c_t, \label{eq:avg} \\
S_c^{\text{min}} &= \min_{t} c_t, \label{eq:min} \\
S_c^{\text{prod}} &= \Bigl(\prod_{t=1}^{T} c_t\Bigr)^{1/T}. \label{eq:product}
\end{align}
The \emph{last} strategy ($S_c^{\text{last}}=c_T$) reports the final-step confidence, which for the methods that propagate uncertainty through history (UAM and the proposed method) already summarizes the trajectory via the aggregation built into the method itself.
The \emph{product} strategy is the geometric mean of the per-step confidences and instantiates the joint-validity estimate that Zhang et al.~\cite{b21} use to formalize the ``Spiral of Hallucination,'' in which a single low-confidence step compounds multiplicatively and collapses the whole-trajectory score.
For request uncertainty $u_1, \ldots, u_T$ we report four strategies:
\begin{align}
S_u^{\text{first}} &= u_1, \label{eq:first} \\
S_u^{\text{avg}} &= \frac{1}{T}\sum_{t=1}^{T} u_t, \label{eq:uavg} \\
S_u^{\text{max}} &= \max_{t} u_t, \label{eq:umax} \\
S_u^{\text{prod}} &= 1 - \Bigl(\prod_{t=1}^{T}(1-u_t)\Bigr)^{1/T}. \label{eq:uproduct}
\end{align}
These four strategies mirror the action-confidence aggregations of Eqs.~(\ref{eq:last})--(\ref{eq:product}) with $1-u_t$ playing the role of the confidence $c_t$: averaging (Eq.~(\ref{eq:uavg})) and the geometric mean (Eq.~(\ref{eq:uproduct}), applied to $\prod_t(1-u_t)$) carry over directly, the conservative aggregation flips from the minimum confidence to the maximum uncertainty (Eq.~(\ref{eq:umax})), and the single-step score is taken at the first step (Eq.~(\ref{eq:first})), where underspecification is typically most apparent, rather than the last.
As we report in Table~\ref{tab:fault_clarification_full} and discuss in Section~\ref{sec:discussion}, the choice of aggregation strategy introduces a substantial hyperparameter that can dominate the effect of the uncertainty method itself.

\section{Experimental Setup}
\label{sec:setup}

We describe the benchmarks used for evaluation (Section~\ref{sec:benchmarks}), the metrics reported (Section~\ref{sec:metrics}), the models and prompt configurations used for data collection (Section~\ref{sec:impl}), and the evaluation protocol that ties them together (Section~\ref{sec:exp-design}).

\subsection{Benchmarks}
\label{sec:benchmarks}

We evaluate on five benchmark configurations, grouped into standard benchmarks for fault detection and clarification-augmented variants for clarification seeking.

\paragraph{Standard benchmarks (fault detection)}
We use three standard interactive benchmarks, with no deliberately injected ambiguity.
On these we evaluate fault detection, the conventional task on which uncertainty estimation methods are evaluated:
\begin{itemize}
    \item \textbf{WebShop}~\cite{b35}: Online shopping over 1{,}000+ products, where the agent searches, filters, and selects items matching natural-language instructions.
    \item \textbf{ALFWorld}~\cite{b36}: Household embodied agent performing tasks (e.g., ``put a clean mug on the desk'') in text-based simulations across 6 task types.
    \item \textbf{REAL}~\cite{b37}: Practical multi-turn tasks on deterministic simulations of 11 real websites.
\end{itemize}

\paragraph{Clarification-augmented variants}
Starting from the standard benchmarks above, we construct two clarification-augmented variants by deliberately underspecifying 50\% of tasks.
This lets us evaluate whether the agent can distinguish underspecified from fully specified goals and trigger \reqclar accordingly:
\begin{itemize}
    \item \textbf{WebShop-Clarification}: Starting from WebShop, for underspecified tasks we strip attribute words (e.g., ``black'', ``leather'') and option clauses (e.g., ``with color: black'') from the instruction, keeping only the base product type and price constraint.
The modification preserves task feasibility (the environment still contains valid products) while removing the information the user would normally provide.
    \item \textbf{ALFWorld-Clarification}: Starting from ALFWorld, for underspecified tasks we randomly remove either the object or the receptacle from the goal (e.g., ``put something in container'' or ``put a mug somewhere'').
\end{itemize}
In both variants, the agent can emit \reqclar to flag the task as ambiguous.
When it does so on an underspecified task, the original fully specified goal is revealed and the episode continues---simulating a user who supplies the missing details---so that task success rate is measured fairly even on tasks where clarification was required.
Each task in these variants carries a binary underspecification label $z$ that records whether it was deliberately underspecified ($z=1$) or left fully specified ($z=0$); this label is the ground truth against which the agent's clarification decision is scored.

We note that evaluation of uncertainty-aware agents remains an open challenge~\cite{b38}; our benchmarks specifically isolate the clarification-seeking capability.

\subsection{Metrics}
\label{sec:metrics}

We report two metric families together with task success rate: the fault-detection metrics from Eqs.~(\ref{eq:roc})--(\ref{eq:ece}) (ROC-AUC, ECE, Brier) and the clarification-seeking metrics from Eqs.~(\ref{eq:precision})--(\ref{eq:acc}) (Precision, Recall, F1, Accuracy).
The fault-detection metrics gauge how well a method solves the conventional task on which uncertainty estimation methods are evaluated---predicting from a trajectory's uncertainty signals whether it will fail---while the clarification-seeking metrics gauge the distinctive capability the proposed decomposition is meant to enable: recognizing an underspecified goal and acting on it.
Fault-detection metrics use the trajectory-level score $S(\tau)$ paired with the success label $y(\tau)$ and are reported per (method, aggregation) pair on all five benchmarks; clarification-seeking metrics use the binary clarification decision $D(\tau)$ paired with the underspecification label $z(\tau)$ and are reported per method on the two clarification-augmented benchmarks.

\subsection{Models and Implementation Details}
\label{sec:impl}

Across the results we report all five backbones (GPT-5.1, DeepSeek-v3.2-exp, GLM-4.7, Qwen3.5-35B, GPT-OSS-120B), so we can check whether the qualitative findings persist across models.
Because the choice of trajectory-level aggregation is critical for fault detection, the fault-detection deep-dive in Section~\ref{sec:res-fault} reports only GPT-5.1 for legibility; the corresponding fault-detection metrics for the remaining four backbones are given in Appendix Table~\ref{tab:fault_standard_full}.
For the proposed method we use $\theta = 0.5$ as the standard clarification threshold and report a dedicated sensitivity ablation separately in Section~\ref{sec:res-threshold}.

\subsection{Evaluation Protocol}
\label{sec:exp-design}

We evaluate each of the three methods (ReAct+UE, UAM, and the proposed method) on each of the five benchmark configurations described in Section~\ref{sec:benchmarks}.
For every task in each benchmark we run the method once with the corresponding prompt instrumentation $\phi$; this yields a trajectory $\tau = (o_1, a_1, \ldots, o_T, a_T)$, a success label $y(\tau)$, and per-step signals $(s_1, \ldots, s_T)$.
Each (method, benchmark) pair is run on 100 tasks.

The trajectory-level continuous score $S(\tau) = \mathrm{Agg}(s_1, \ldots, s_T)$, computed for each aggregation in Section~\ref{sec:method-agg}, is paired with the success label $y(\tau)$ to produce the fault-detection metrics of Eqs.~(\ref{eq:roc})--(\ref{eq:ece}); these are computed on all five benchmarks, since every task carries a success label.
The binary clarification decision $D(\tau)$ of Section~\ref{sec:ps-clarif} is paired with the underspecification label $z(\tau)$ to produce the clarification-seeking metrics of Eqs.~(\ref{eq:precision})--(\ref{eq:acc}); these are computed only on the two clarification-augmented variants.

\section{Results}
\label{sec:results}

We present the quantitative results, organized around the two task families: Sections~\ref{sec:res-clar} and~\ref{sec:res-fault} report the headline results on the clarification-augmented and standard benchmarks respectively.
Section~\ref{sec:res-sr} then consolidates the task success-rate trend that spans both benchmark families, and Section~\ref{sec:res-calib} reports the calibration of the per-step confidence signals.
Section~\ref{sec:res-length} presents a diagnostic showing that product aggregation largely tracks trajectory length, which informs how we interpret its fault-detection scores; Section~\ref{sec:res-threshold} reports a sensitivity analysis over the clarification threshold $\theta$.

\subsection{Clarification Seeking on Modified Benchmarks}
\label{sec:res-clar}

\begin{figure*}[!t]
\centering
\includegraphics[width=0.95\textwidth]{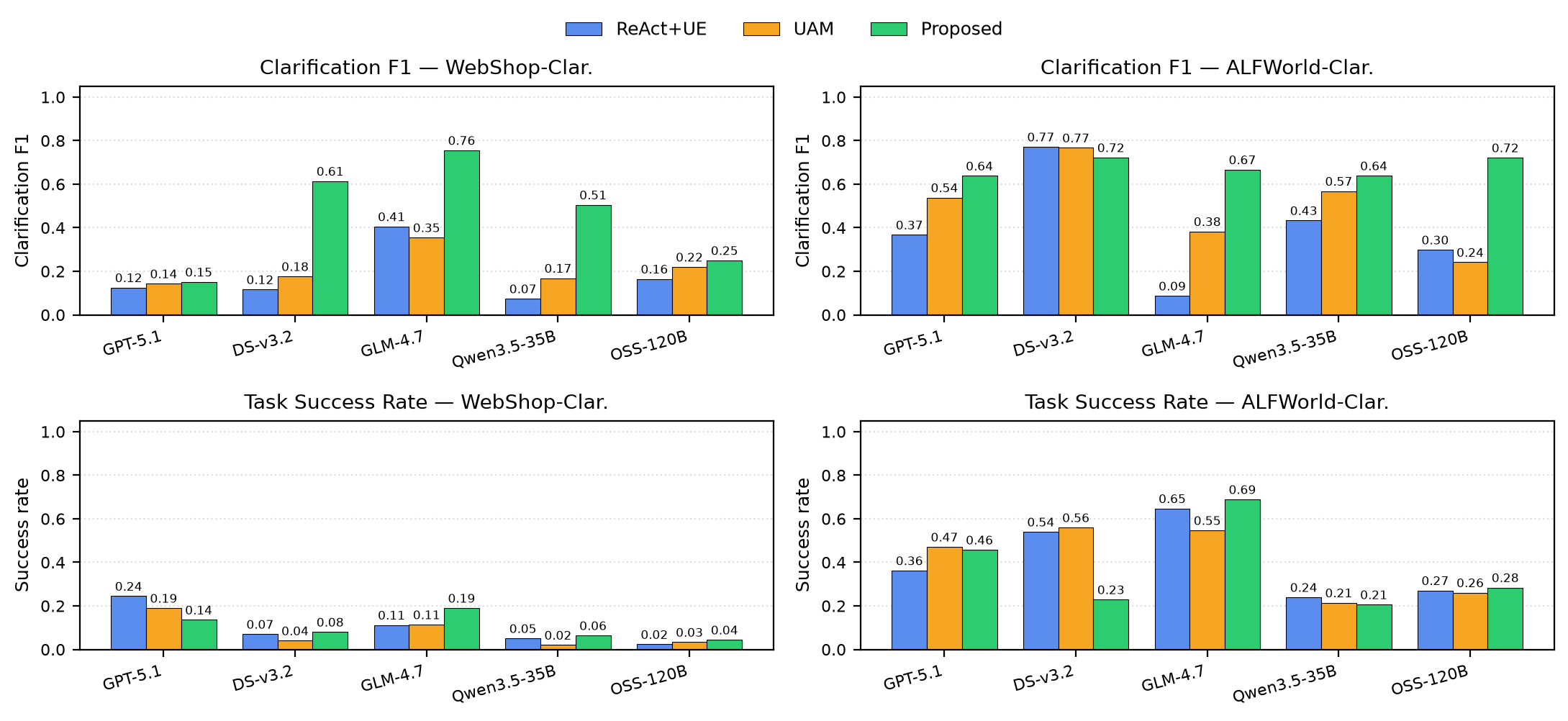}
\caption{Clarification F1 (top) and task success rate (bottom) on the two clarification-augmented benchmarks across all five LLM backbones.
Bars are grouped by method.
The proposed method leads clarification F1 on every backbone on WebShop-Clar.\ and on four of five on ALFWorld-Clar., confirming that the decomposition enables clarification seeking where scalar-confidence baselines cannot.}
\label{fig:clarification_bars}
\end{figure*}

Figure~\ref{fig:clarification_bars} reports clarification F1 and task success rate across all backbones on both clarification-augmented benchmarks.
The proposed method leads F1 on most (backbone, benchmark) pairs -- in particular on every backbone on WebShop-Clar.\ and on four of five on ALFWorld-Clar. -- leading by roughly 0.28 in absolute F1 over ReAct+UE and 0.21 over UAM when averaged across the five backbones on each benchmark, confirming that the method generalizes beyond a single backbone LLM.

\subsection{Fault Detection on Standard Benchmarks}
\label{sec:res-fault}

\begin{figure*}[!t]
\centering
\includegraphics[width=0.95\textwidth]{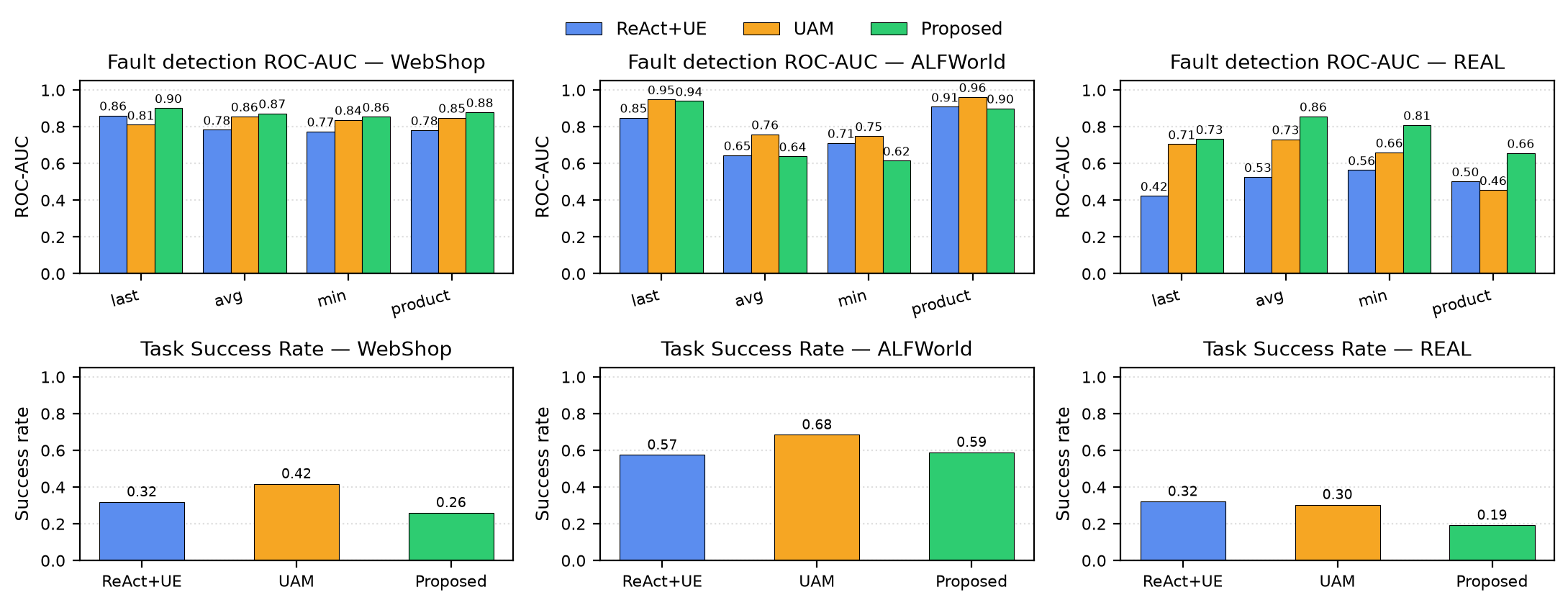}
\caption{Fault-detection ROC-AUC across the four trajectory-level aggregations (top) and task success rate (bottom) on the three standard benchmarks (GPT-5.1).
Bars are grouped by method.
Across aggregations and benchmarks the three methods achieve comparable fault-detection ROC-AUC, confirming that adding the request-uncertainty decomposition does not sacrifice the conventional uncertainty objective.}
\label{fig:fault_detection_bars}
\end{figure*}

Figure~\ref{fig:fault_detection_bars} reports fault-detection ROC-AUC under all four aggregations, together with task success rate, on the three standard benchmarks for GPT-5.1.
The proposed method preserves discrimination: it reaches the highest \emph{last}- and \emph{avg}-aggregation ROC-AUC on WebShop and REAL and remains within 0.08 of the best ROC-AUC on ALFWorld.
On ALFWorld, product aggregation attains the highest fault-detection ROC-AUC of any aggregation--benchmark combination; we quantify this effect in Section~\ref{sec:res-length} and discuss its implications in Section~\ref{sec:disc-agg}.

The decomposition does not come at the cost of fault detection: across backbones the proposed method solves this conventional uncertainty task on par with the ReAct+UE and UAM baselines, with full per-(backbone, method, aggregation) metrics in Appendix Table~\ref{tab:fault_standard_full}.

\subsection{Task Success Rate Across Methods}
\label{sec:res-sr}

The success-rate panels of Figures~\ref{fig:clarification_bars} and~\ref{fig:fault_detection_bars} show that task success rate tends to decline as the agent is given more uncertainty instrumentation.
Averaged across all five benchmarks and all five backbones, mean success rate falls monotonically from $28.6\%$ for ReAct+UE to $27.8\%$ for UAM and $27.0\%$ for the proposed method; success rate for every backbone and method is reported in Appendix Tables~\ref{tab:fault_standard_full} and~\ref{tab:clarification_full}.
We call this effect \emph{capability dilution} and examine it in Section~\ref{sec:disc-dilution}.

\subsection{Calibration}
\label{sec:res-calib}

\begin{figure*}[!t]
\centering
\includegraphics[width=0.95\textwidth]{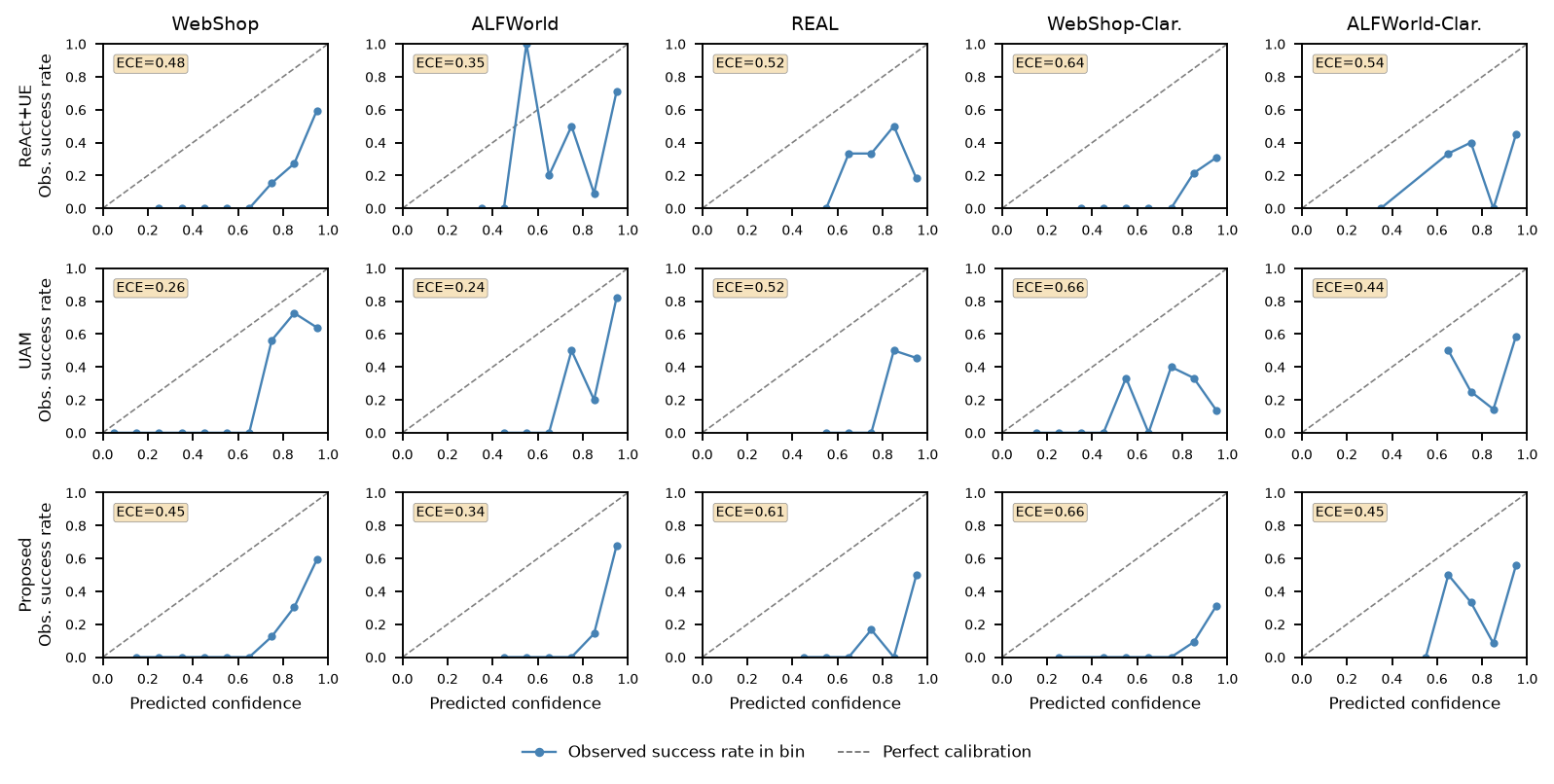}
\caption{Reliability diagrams for GPT-5.1: the three methods (rows) across the five benchmarks (columns), under last-step aggregation.
Each point bins trajectories by action confidence and plots observed success rate against the bin's predicted confidence; the dashed line marks perfect calibration.
All curves lie below the diagonal, indicating systematic overconfidence for every method and benchmark.}
\label{fig:calibration-gpt51}
\end{figure*}

Figure~\ref{fig:calibration-gpt51} reports reliability diagrams for the three methods across all five benchmarks for GPT-5.1.
Across every method and benchmark the curve lies below the diagonal---predicted confidence systematically exceeds observed success rate---with per-panel ECE ranging from 0.24 to 0.66.
We examine the implications of this overconfidence in Section~\ref{sec:disc-calib}; the corresponding diagrams for the remaining four backbones are given in Appendix~\ref{app:calibration}.

\subsection{Product Aggregation and Trajectory Length}
\label{sec:res-length}

Table~\ref{tab:length_proxy} reports fault-detection metrics for trajectories in which the real per-step confidences are replaced by two confidence-free surrogates---i.i.d.\ $\mathcal{U}(0,1)$ draws of matched trajectory length (averaged over 50 seeds) and the deterministic $1/\text{length}$ score---with product aggregation re-applied to each.
On ALFWorld both surrogates reach high ROC-AUC (0.92--0.99), matching or exceeding the real-product score for every method---with the lone exception of the random surrogate under UAM (0.94 vs.\ 0.96); the effect is weaker but present on WebShop and small on REAL.
We discuss what this length confound implies for interpreting product-aggregation scores in Section~\ref{sec:disc-agg}.

\begin{table*}[t]
\centering
\caption{Product aggregation as a length proxy on the standard benchmarks. \emph{random} draws per-step values from $\mathcal{U}(0,1)$ matched to each trajectory length, averaged over 50 seeds; $1/\text{length}$ uses the trajectory step count directly. Best per (method, benchmark, metric) in bold (GPT-5.1). On ALFWorld both confidence-free surrogates reach ROC-AUC of 0.92--0.99, matching or exceeding the real-product score for nearly every method (the lone exception being the random surrogate under UAM), confirming that high product-aggregation ROC-AUC on that benchmark reflects trajectory-length confounding rather than an informative confidence signal.}
\label{tab:length_proxy}
\setlength{\tabcolsep}{3pt}
\footnotesize
\begin{tabular}{|l|l|ccc|ccc|ccc|}
\hline
 & & \multicolumn{3}{c|}{\textbf{WebShop}} & \multicolumn{3}{c|}{\textbf{ALFWorld}} & \multicolumn{3}{c|}{\textbf{REAL}} \\
\textbf{Method} & \textbf{Variant} & ROC-AUC & ECE & Brier & ROC-AUC & ECE & Brier & ROC-AUC & ECE & Brier \\
\hline
\multirow{3}{*}{ReAct+UE} & real product & \textbf{0.781} & \textbf{0.171} & \textbf{0.203} & 0.909 & \textbf{0.378} & \textbf{0.300} & 0.503 & \textbf{0.357} & \textbf{0.347} \\
 & random ($n$=50) & 0.712 & 0.312 & 0.311 & 0.916 & 0.565 & 0.557 & \textbf{0.540} & 0.384 & 0.361 \\
 & $1/\text{length}$ & 0.735 & 0.207 & 0.242 & \textbf{0.957} & 0.485 & 0.443 & 0.529 & 0.396 & 0.401 \\
\hline
\multirow{3}{*}{UAM} & real product & \textbf{0.849} & \textbf{0.317} & \textbf{0.275} & 0.962 & \textbf{0.441} & \textbf{0.327} & \textbf{0.455} & \textbf{0.226} & \textbf{0.296} \\
 & random ($n$=50) & 0.792 & 0.411 & 0.406 & 0.944 & 0.675 & 0.667 & 0.395 & 0.382 & 0.352 \\
 & $1/\text{length}$ & 0.827 & 0.320 & 0.314 & \textbf{0.976} & 0.586 & 0.527 & 0.375 & 0.361 & 0.408 \\
\hline
\multirow{3}{*}{Proposed} & real product & \textbf{0.879} & \textbf{0.117} & \textbf{0.143} & 0.900 & \textbf{0.328} & \textbf{0.255} & \textbf{0.657} & \textbf{0.212} & \textbf{0.209} \\
 & random ($n$=50) & 0.814 & 0.254 & 0.251 & 0.940 & 0.581 & 0.574 & 0.521 & 0.266 & 0.238 \\
 & $1/\text{length}$ & 0.835 & 0.162 & 0.192 & \textbf{0.991} & 0.497 & 0.451 & 0.514 & 0.296 & 0.316 \\
\hline
\end{tabular}
\end{table*}

\subsection{Clarification Threshold Sensitivity}
\label{sec:res-threshold}

Table~\ref{tab:threshold_xmodel} reports clarification-seeking metrics for the proposed method under three values of the clarification threshold, $\theta \in \{0.25, 0.5, 0.75\}$, across all five backbones on both clarification-augmented benchmarks.

\begin{table*}[t]
\centering
\caption{Clarification-threshold sensitivity for the proposed method across all five backbones. Best per (backbone, benchmark, metric) in bold across the three $\theta$ rows. Averaged across all five backbones, $\theta = 0.25$ attains the highest mean clarification F1 on both benchmarks (by at most 0.03 over $\theta = 0.5$), though no single $\theta$ dominates every (model, benchmark) pair; we report $\theta = 0.5$ as a balanced default in the main results.}
\label{tab:threshold_xmodel}
\setlength{\tabcolsep}{3pt}
\footnotesize
\resizebox{\textwidth}{!}{%
\begin{tabular}{|l|c|ccccc|ccccc|}
\hline
 & & \multicolumn{5}{c|}{\textbf{WebShop-Clar.}} & \multicolumn{5}{c|}{\textbf{ALFWorld-Clar.}} \\
\textbf{Model} & $\theta$ & Success Rate (\%) & Precision & Recall & F1 & Accuracy & Success Rate (\%) & Precision & Recall & F1 & Accuracy \\
\hline
\multirow{3}{*}{GPT-5.1} & 0.25 & \textbf{14.0} & \textbf{0.375} & \textbf{0.125} & \textbf{0.188} & 0.441 & 44.1 & 0.900 & \textbf{0.621} & \textbf{0.735} & \textbf{0.809} \\
 & 0.50 & 13.6 & 0.267 & 0.105 & 0.151 & \textbf{0.444} & \textbf{45.7} & 0.938 & 0.484 & 0.638 & 0.757 \\
 & 0.75 & 7.6 & 0.300 & 0.071 & 0.115 & 0.418 & 31.0 & \textbf{0.989} & 0.037 & 0.071 & 0.552 \\
\hline
\multirow{3}{*}{DeepSeek-v3.2-exp} & 0.25 & 5.0 & 0.509 & 0.580 & 0.542 & 0.510 & 13.8 & 0.554 & 0.976 & 0.707 & 0.609 \\
 & 0.50 & \textbf{8.0} & \textbf{0.547} & \textbf{0.700} & \textbf{0.614} & \textbf{0.560} & 22.9 & 0.591 & 0.929 & 0.722 & 0.639 \\
 & 0.75 & 3.0 & 0.304 & 0.140 & 0.192 & 0.410 & \textbf{34.8} & \textbf{0.796} & \textbf{0.977} & \textbf{0.878} & \textbf{0.865} \\
\hline
\multirow{3}{*}{GLM-4.7} & 0.25 & 17.1 & 0.523 & 0.639 & 0.575 & 0.514 & 56.0 & 0.727 & \textbf{0.640} & \textbf{0.681} & \textbf{0.700} \\
 & 0.50 & \textbf{19.0} & \textbf{0.694} & \textbf{0.829} & \textbf{0.756} & \textbf{0.722} & \textbf{68.9} & 0.737 & 0.609 & 0.667 & 0.689 \\
 & 0.75 & 17.6 & 0.548 & 0.486 & 0.515 & 0.568 & 57.4 & \textbf{0.990} & 0.296 & 0.457 & 0.648 \\
\hline
\multirow{3}{*}{Qwen3.5-35B} & 0.25 & \textbf{10.3} & \textbf{0.633} & \textbf{0.633} & \textbf{0.633} & \textbf{0.629} & 24.2 & \textbf{0.611} & \textbf{0.815} & \textbf{0.698} & \textbf{0.694} \\
 & 0.50 & 6.4 & 0.500 & 0.511 & 0.505 & 0.500 & 20.6 & 0.561 & 0.742 & 0.639 & 0.618 \\
 & 0.75 & 5.3 & 0.388 & 0.404 & 0.396 & 0.389 & \textbf{27.5} & 0.545 & 0.462 & 0.500 & 0.652 \\
\hline
\multirow{3}{*}{GPT-OSS-120B} & 0.25 & 3.3 & \textbf{0.339} & \textbf{0.447} & \textbf{0.385} & 0.264 & 17.2 & 0.611 & \textbf{0.917} & \textbf{0.733} & 0.656 \\
 & 0.50 & \textbf{4.4} & 0.245 & 0.255 & 0.250 & 0.209 & \textbf{28.4} & 0.714 & 0.729 & 0.722 & \textbf{0.716} \\
 & 0.75 & 2.1 & 0.250 & 0.224 & 0.237 & \textbf{0.268} & 25.8 & \textbf{0.714} & 0.500 & 0.588 & 0.639 \\
\hline
\end{tabular}}
\end{table*}

No single $\theta$ dominates every (model, benchmark) pair in Table~\ref{tab:threshold_xmodel}; averaged across all five backbones, the lowest threshold $\theta = 0.25$ attains the highest mean clarification F1 on both benchmarks (WebShop-Clar.: 0.464, vs.\ 0.455 at $\theta = 0.5$ and 0.291 at $\theta = 0.75$; ALFWorld-Clar.: 0.71, vs.\ 0.68 at $\theta = 0.5$ and 0.50 at $\theta = 0.75$), though $\theta = 0.25$ and $\theta = 0.5$ differ by at most 0.03; we report $\theta = 0.5$ as a balanced default in the main results (Table~\ref{tab:clarification_full}).
We discuss why we threshold a scalar uncertainty and where this leaves the method in Section~\ref{sec:native}.

\section{Discussion}
\label{sec:discussion}

Section~\ref{sec:disc-decomp} first asks what kind of signal the decomposition actually captures.
Section~\ref{sec:disc-dilution} then analyzes how extending the agent's prompt with additional uncertainty objectives affects task success rate.
Section~\ref{sec:disc-calib} examines the overconfidence pattern surfaced by the calibration results of Section~\ref{sec:res-calib}.
Section~\ref{sec:disc-agg} explains why the aggregation choice can dominate the method choice, using the length-proxy finding from Table~\ref{tab:length_proxy}.
Finally, Section~\ref{sec:native} argues that these patterns motivate moving uncertainty estimation from the prompt into the model itself.

\subsection{Why Decomposition Helps}
\label{sec:disc-decomp}

The key advantage of separating request uncertainty from action confidence is that it gives the agent a dedicated channel for expressing goal ambiguity.
With a single confidence score, an agent that encounters an underspecified task must either (a) report low confidence on its actions, which is indistinguishable from genuinely difficult but fully specified tasks, or (b) proceed with an arbitrary interpretation and report high confidence, which leads to silent failure.
The decomposition resolves this ambiguity, directly addressing the call by Kirchhof et al.~\cite{b13} for ``underspecification uncertainty'' as a distinct category.

This aligns with the broader agentic interpretability vision of Kim et al.~\cite{b14}: by explicitly communicating why it is uncertain (task difficulty vs.\ goal ambiguity), the agent helps users build a mental model of its reasoning, rather than presenting an opaque scalar score.

\subsection{Capability Dilution}
\label{sec:disc-dilution}

Inspection of trajectories where the proposed method fails but ReAct+UE succeeds reveals a recurring pattern: the proposed agent spends a disproportionate share of its reasoning budget debating whether the goal is underspecified, producing a long $u_t$ explanation and only a perfunctory action rationale.
This is consistent with a bounded reasoning budget that must be split across task-solving, confidence estimation, and request-uncertainty assessment.
The proposed method asks the LLM to do all three in a single forward pass with no additional tokens of reasoning allocated to compensate.
The monotonic drop from UAM ($27.8\%$) to the proposed method ($27.0\%$) isolates this effect: since the two methods differ only in the addition of the request-uncertainty signal $u_t$ and its explanation $x_t$, the $0.8$ percentage-point SR gap is directly attributable to the enlarged prompt rather than to any other architectural change.

We call this \emph{capability dilution}: each additional uncertainty objective added to the prompt degrades the primary task objective.
The degradation is concentrated on benchmarks where the task itself is already demanding (REAL: $32.1\% \to 19.2\%$; WebShop: $31.9\% \to 26.1\%$) and is absent on the more structured ALFWorld, where success rate is essentially unchanged ($57.5\% \to 58.9\%$).
This is a limitation that cannot be resolved by better prompt engineering within the prompt-only regime, because the root cause is a fixed shared reasoning budget rather than a specific phrasing failure.

\subsection{Overconfidence and Calibration}
\label{sec:disc-calib}

The calibration results of Section~\ref{sec:res-calib} (with the full per-backbone reliability diagrams in Appendix~\ref{app:calibration}) show that all three methods sit well below the diagonal: predicted confidence consistently exceeds observed success rate.
We interpret this as a structural bias of prompt-based self-reported confidence -- an agent that has already committed to an action has an incentive to justify rather than critique it -- rather than an issue specific to any one method.
This matches the independent findings of Kaddour et al.~\cite{b30} and the survey observation by Oh et al.~\cite{b17} that ``dynamically expanding context memory results in increasingly inflated and unreliable verbalized confidence.'' The practical implication is that while the confidence scores are useful as \emph{ranking} signals (ROC-AUC), they cannot be interpreted as probabilities without a post-hoc recalibration step.
Evaluation methodology itself introduces additional uncertainty: the choice of correctness function can substantially affect UQ method rankings~\cite{b39}.

\subsection{Aggregation as Hidden Hyperparameter}
\label{sec:disc-agg}

Across Appendix Tables~\ref{tab:fault_standard_full} and~\ref{tab:fault_clarification_full}, the best aggregation strategy differs by method and benchmark: product excels on ALFWorld, avg excels on REAL for both UAM and the proposed method, and avg/last excel on WebShop for the proposed method.
In practice this means a practitioner tuning only the aggregation can produce arbitrarily large differences between methods without changing the underlying uncertainty signal.
This matches Oh et al.'s~\cite{b17} observation that naive cascade aggregations cannot robustly distinguish successful from failed trajectories.

Product aggregation (Eq.~(\ref{eq:product})) was introduced as an operationalization of the ``Spiral of Hallucination'' formalized by Zhang et al.~\cite{b21}.
The length-proxy experiment of Table~\ref{tab:length_proxy} shows, however, that on ALFWorld this aggregation does not capture that mechanism but instead behaves as a trajectory-length proxy: replacing real confidences with i.i.d.\ $\mathcal{U}(0,1)$ draws, or with a deterministic $1/\text{length}$ value, matches or exceeds the real-product ROC-AUC for every method.
Failed ALFWorld tasks are systematically longer than successful ones, and the geometric mean of $T$ values in $[0,1]$ decreases with $T$, so the length signal dominates.
The effect is weaker but present on WebShop and essentially absent on REAL.
The qualitative takeaway is that high product-aggregation ROC-AUC should not be interpreted as evidence that the confidence signal is informative; it can simply be evidence that the agent took more steps to fail.

\subsection{Toward Native Uncertainty Estimation}
\label{sec:native}

The limitations identified above -- capability dilution, systematic overconfidence, and aggregation sensitivity -- are not specific to the proposed method but are fundamental to the prompt-based paradigm.
Prompt-based methods bolt uncertainty estimation onto a model that was not designed for it: the model must simultaneously solve the task and accurately assess its own confidence, competing objectives that share a fixed reasoning budget.

To compare the decomposition against the scalar-confidence baselines on equal footing, we have the agent emit a numeric request-uncertainty value and route on a fixed threshold, which is what the ablation of Section~\ref{sec:res-threshold} probes.
We also agree with the position papers that motivate this work that the more promising direction is to move beyond such scalar thresholds toward proactive interaction grounded in human-readable, interpretable explanations of the agent's uncertainty~\cite{b13},~\cite{b14}.

Training-based approaches offer a compelling alternative.
Suri et al.~\cite{b29} demonstrate this with SAGE-Agent, where GRPO-based fine-tuning improved a 3B-parameter model's clarification accuracy from 36.5\% to 65.2\% -- a 78.7\% relative improvement.
Their certainty-weighted reward function aligns the model to produce calibrated uncertainty estimates natively, eliminating the need for prompt-based elicitation and the associated capability dilution.
Similarly, Chen et al.~\cite{b34} show that uncertainty-aware self-training for GUI agents produces better-calibrated confidence estimates than prompting alone.

We argue that the most promising path toward practical agentic uncertainty lies in alignment-based approaches that natively integrate uncertainty estimation, decomposition, and addressing (clarification, abstention) into the model itself. Specifically:
\begin{itemize}
    \item \textbf{Native decomposition.} Rather than prompting for separate $c_t$ and $u_t$ scores, models should be trained (via RLHF, DPO, or GRPO) to internally distinguish between task difficulty and goal ambiguity, producing decomposed uncertainty as a natural part of their output.
    \item \textbf{Integrated aggregation.} Instead of applying post-hoc aggregation strategies, models should learn to maintain and propagate trajectory-level uncertainty internally, eliminating the aggregation hyperparameter entirely.
    \item \textbf{Calibrated communication.} Following Kirchhof et al.'s~\cite{b13} call for rich output uncertainties, aligned models should communicate uncertainty in natural language -- explaining what is uncertain and why -- rather than producing poorly calibrated scalar scores.
\end{itemize}
This vision aligns with the agentic interpretability paradigm of Kim et al.~\cite{b14}, where agents proactively build shared understanding with users.
The proposed prompt-based decomposition demonstrates that the concept of separating request uncertainty from action confidence is effective; the challenge now is to move this decomposition from the prompt into the model itself.

\section{Limitations}
\label{sec:limitations}

The analysis exposes three limitations of the prompt-based paradigm that motivated this work.
First, adding uncertainty instructions consistently degrades task-solving ability (\emph{capability dilution}, Section~\ref{sec:disc-dilution}).
Second, all methods suffer from systematic overconfidence visible in the calibration plots (Section~\ref{sec:disc-calib}).
Third, trajectory-level aggregation introduces a consequential hyperparameter, and its product variant can act as a trajectory-length proxy rather than a confidence signal (Section~\ref{sec:disc-agg}).
Together these limit the regime in which scalar prompt-based confidence can be relied upon.

\section{Future Work}
\label{sec:future}

The clarification-augmented benchmarks rely on synthetic underspecification -- stripping attributes or objects -- and score clarification as a single binary action.
Future evaluations should curate tasks in which ambiguity is organic rather than procedurally generated, and assess the linguistic quality and informativeness of the agent's clarifying questions, following the methodology of SAGE-Agent~\cite{b29}.
The decomposition itself should also move out of the prompt and into the model, as argued in Section~\ref{sec:native}; operationalizing it via RLHF, DPO, or GRPO would test whether its benefits survive once capability dilution is removed as a confound.

\section{Conclusion}
\label{sec:conclusion}

We presented a prompt-based decomposition of agentic uncertainty into action confidence and request uncertainty, enabling proactive clarification seeking by giving the agent a dedicated channel for goal ambiguity that a single confidence score conflates with task difficulty.
Alongside the method, we contributed two clarification-augmented benchmarks (WebShop-Clarification and ALFWorld-Clarification) and a systematic comparison of the prompt-based family (ReAct+UE, UAM, and the proposed method) across five LLM backbones.
The proposed method leads on the clarification-augmented benchmarks across multiple backbones, while the limitations summarized above show that prompt-based methods are best viewed as proofs of concept and that the next step is to move the decomposition out of the prompt and into the model via alignment-based training.

\section*{Acknowledgments}

The author thanks Danil Silantyev (NDDev, Kazakhstan; ITMO University, St.\ Petersburg, Russia) for his help with the code and for financially supporting the experiments.

\FloatBarrier
\appendix
\makeatletter
\renewcommand\@seccntformat[1]{%
  \ifnum\pdfstrcmp{#1}{section}=0
    \appendixname~\csname the#1\endcsname:\space
  \else
    \csname the#1\endcsname\quad
  \fi
}
\makeatother
\section{Prompts}
\label{app:prompts}

Each prompt is a system message and a user message rebuilt at every step.
The user message is a sequence of blocks: \emph{Goal}, \emph{Observation}, \emph{Action Space} (runtime context); optional \emph{History} and \emph{Error}; \emph{Next-Action} instruction; \emph{Confidence Elicitation} suffix.
ALFWorld concatenates these into one prose block.
Below, for each method and benchmark, we list the static instructional blocks (system message, next-action / action-selection block, history-entry template, confidence elicitation suffix); runtime-only context blocks are not shown.

\subsection{ReAct+UE}
\label{app:prompt-react}

\subsubsection{WebShop}
\label{app:prompt-react-webshop}

\paragraph{System message}

\begin{quote}\small\itshape\raggedright
You are a shopping agent.\ Your goal is to find and buy a product that matches the given instruction on a simulated web store.\\[2pt]
Available actions:\\
\hspace*{1em}\texttt{search[keywords]}  --  search for products using keywords.\\
\hspace*{1em}\texttt{click[value]}  --  click a button or link; value must exactly match one of the available clickables listed in the observation.\\
\hspace*{1em}\texttt{request\_clarification}  --  request a more specified goal if the request is missing key details or has multiple valid solutions (e.g.\ color, size).\\[2pt]
Output format (required):\\
\hspace*{1em}\texttt{<think>...</think>}\\
\hspace*{1em}\texttt{<action>search[\ldots] or click[\ldots] or request\_clarification</action>}\\
\hspace*{1em}\texttt{<confidence>0.0--1.0</confidence>}\\
\hspace*{1em}\texttt{<explanation>...</explanation>}
\end{quote}

\paragraph{Next-action block}

\begin{quote}\small\itshape\raggedright
You are now at step \{t\}.\ Prior to this step, you have already taken \{t\} step(s).\ Now it's your turn to take an action.\\[2pt]
If the goal is ambiguous or missing key details, you should seek clarification before acting.
\end{quote}

\paragraph{History-entry template}

\begin{quote}\small\itshape\raggedright
Step \{i\}:\ Observation: \{$o_i$\}\\
Action: \texttt{<think>}\{$r_i$\}\texttt{</think>} \texttt{<action>}\{$a_i$\}\texttt{</action>}
\end{quote}

\paragraph{Confidence elicitation suffix}

\begin{quote}\small\itshape\raggedright
After your action, you MUST provide:\\[2pt]
1.\ Your confidence level (0.0--1.0) in \texttt{<confidence>...</confidence>} tags.\\[2pt]
2.\ An explanation of your confidence in \texttt{<explanation>...</explanation>} tags:\\
 --  Explain what makes you confident.\\
 --  Explain what concerns or uncertainties you have.\\
 --  What information might be missing or unclear.\\
 --  What alternative actions you considered.\\
 --  DO NOT output empty \texttt{<explanation></explanation>} tags  --  you MUST provide actual text inside.
\end{quote}

\subsubsection{ALFWorld}
\label{app:prompt-react-alfworld}

\paragraph{System message}

\begin{quote}\small\itshape\raggedright
You are an expert agent operating in the ALFRED Embodied Environment.
\end{quote}

\paragraph{Action-selection block}

\begin{quote}\small\itshape\raggedright
Now it's your turn to take an action.\\
You should first reason step-by-step about the current situation.\ This reasoning process MUST be enclosed within \texttt{<think> </think>} tags.\\
Once you've finished your reasoning, you should choose an admissible action for the current step and present it within \texttt{<action> </action>} tags.\\[2pt]
If the goal is ambiguous or missing key details, you should seek clarification before acting.
\end{quote}

\paragraph{History-entry template} Identical to Appendix~\ref{app:prompt-react-webshop}.

\paragraph{Confidence elicitation suffix} Identical to Appendix~\ref{app:prompt-react-webshop}.

\subsubsection{REAL}
\label{app:prompt-react-real}

\paragraph{System message}

\begin{quote}\small\itshape\raggedright
\# Instructions\\[2pt]
Review the current state of the page and all other information to find the best possible next action to accomplish your goal.\ Your answer will be interpreted and executed by a program, make sure to follow the formatting instructions.\\[2pt]
You should first reason step-by-step about the current situation.\ This reasoning process MUST be enclosed within \texttt{<think> </think>} tags.\\
Once you've finished your reasoning, you should choose an admissible action for the current step and present it within \texttt{<action> </action>} tags.
\end{quote}

\paragraph{Next-action block}

\begin{quote}\small\itshape\raggedright
You are now at step \{t\}.\ Prior to this step, you have already taken \{t\} step(s).\ Now it's your turn to take an action.
\end{quote}

\paragraph{History-entry template} Identical to Appendix~\ref{app:prompt-react-webshop}.

\paragraph{Confidence elicitation suffix} Identical to Appendix~\ref{app:prompt-react-webshop}.

\subsection{UAM}
\label{app:prompt-uam}

\subsubsection{WebShop}
\label{app:prompt-uam-webshop}

\paragraph{System message} Identical to Appendix~\ref{app:prompt-react-webshop}.

\paragraph{Next-action block} Identical to Appendix~\ref{app:prompt-react-webshop}.

\paragraph{History-entry template}

\begin{quote}\small\itshape\raggedright
Step \{i\}:\ Observation: \{$o_i$\}\\
Action: \texttt{<think>}\{$r_i$\}\texttt{</think>} \texttt{<action>}\{$a_i$\}\texttt{</action>}\\
\texttt{<confidence>}\{$c_i$\}\texttt{</confidence>}\\
\texttt{<explanation>}\{$e_i$\}\texttt{</explanation>}
\end{quote}

\paragraph{Confidence elicitation suffix} Identical to Appendix~\ref{app:prompt-react-webshop}.

\subsubsection{ALFWorld}
\label{app:prompt-uam-alfworld}

\paragraph{System message} Identical to Appendix~\ref{app:prompt-react-alfworld}.

\paragraph{Action-selection block} Identical to Appendix~\ref{app:prompt-react-alfworld}.

\paragraph{History-entry template} Identical to Appendix~\ref{app:prompt-uam-webshop}.

\paragraph{Confidence elicitation suffix} Identical to Appendix~\ref{app:prompt-react-webshop}.

\subsubsection{REAL}
\label{app:prompt-uam-real}

\paragraph{System message} Identical to Appendix~\ref{app:prompt-react-real}.

\paragraph{Next-action block} Identical to Appendix~\ref{app:prompt-react-real}.

\paragraph{History-entry template} Identical to Appendix~\ref{app:prompt-uam-webshop}.

\paragraph{Confidence elicitation suffix} Identical to Appendix~\ref{app:prompt-react-webshop}.

\subsection{Proposed Method}
\label{app:prompt-proposed}

\subsubsection{WebShop}
\label{app:prompt-proposed-webshop}

\paragraph{System message}

\begin{quote}\small\itshape\raggedright
You are a shopping agent.\ Your goal is to find and buy a product that matches the given instruction on a simulated web store.\\[2pt]
Available actions:\\
\hspace*{1em}\texttt{search[keywords]}  --  search for products using keywords.\\
\hspace*{1em}\texttt{click[value]}  --  click a button or link; value must exactly match one of the available clickables listed in the observation.\\
\hspace*{1em}\texttt{request\_clarification}  --  request a more specified goal if the request is missing key details or has multiple valid solutions (e.g.\ color, size).\\[2pt]
Output format (required):\\
\hspace*{1em}\texttt{<think>...</think>}\\
\hspace*{1em}\texttt{<u\_request>0.0--1.0</u\_request>}\\
\hspace*{1em}\texttt{<u\_request\_}\allowbreak\texttt{explanation>...</u\_request\_}\allowbreak\texttt{explanation>}\\
\hspace*{1em}\texttt{<action>search[\ldots] or click[\ldots] or request\_clarification</action>}\\
\hspace*{1em}\texttt{<confidence>0.0--1.0</confidence>}\\
\hspace*{1em}\texttt{<explanation>...</explanation>}
\end{quote}

\paragraph{Next-action block}

\begin{quote}\small\itshape\raggedright
You are now at step \{t\}.\ Prior to this step, you have already taken \{t\} step(s).\ Now it's your turn to take an action.\\[2pt]
After thinking, you MUST assess your request uncertainty (0.0--1.0) in \texttt{<u\_request>...</u\_request>} tags.\\
\hspace*{1em}0.0 = the goal fully specifies every parameter  --  there is exactly one correct solution.\\
\hspace*{1em}0.5 = the goal leaves open at least one choice where the user likely has a specific preference they did not state  --  you would be guessing on their behalf.\\
\hspace*{1em}1.0 = critical details are missing, many equally valid interpretations exist.\\[2pt]
Be meticulous: if the goal leaves ANY parameter open-ended, ask yourself  --  would a real user genuinely be satisfied with ANY valid option, or do they most likely have a specific intent they failed to communicate?\ If you find yourself choosing one option among several equally plausible alternatives without a clear basis, that is a sign \texttt{u\_request} should be high.\\[2pt]
Then explain your assessment in \texttt{<u\_request\_}\allowbreak\texttt{explanation>}\,\ldots\,\texttt{</u\_request\_}\allowbreak\texttt{explanation>} tags.\\[2pt]
If \texttt{u\_request >= }$\theta$, your action MUST be \texttt{request\_clarification}.
\end{quote}

\paragraph{History-entry template}

\begin{quote}\small\itshape\raggedright
Step \{i\}:\ Observation: \{$o_i$\}\\
\texttt{<think>}\{$r_i$\}\texttt{</think>}\\
\texttt{<u\_request>}\{$u_i$\}\texttt{</u\_request>}\\
\texttt{<u\_request\_}\allowbreak\texttt{explanation>}\{$x_i$\}\texttt{</u\_request\_}\allowbreak\texttt{explanation>}\\
\texttt{<action>}\{$a_i$\}\texttt{</action>}\\
\texttt{<confidence>}\{$c_i$\}\texttt{</confidence>}\\
\texttt{<explanation>}\{$e_i$\}\texttt{</explanation>}
\end{quote}

\paragraph{Confidence elicitation suffix} Identical to Appendix~\ref{app:prompt-react-webshop}.

\subsubsection{ALFWorld}
\label{app:prompt-proposed-alfworld}

\paragraph{System message} Identical to Appendix~\ref{app:prompt-react-alfworld}.

\paragraph{Action-selection block}

\begin{quote}\small\itshape\raggedright
Now it's your turn to take an action.\\
You should first reason step-by-step about the current situation.\ This reasoning process MUST be enclosed within \texttt{<think> </think>} tags.\\[2pt]
After thinking, you MUST assess your request uncertainty (0.0--1.0) in \texttt{<u\_request>...</u\_request>} tags.\\
\hspace*{1em}0.0 = the goal fully specifies every parameter  --  there is exactly one correct solution.\\
\hspace*{1em}0.5 = the goal leaves open at least one choice where the user likely has a specific preference they did not state  --  you would be guessing on their behalf.\\
\hspace*{1em}1.0 = critical details are missing, many equally valid interpretations exist.\\[2pt]
Be meticulous: if the goal leaves ANY parameter open-ended, ask yourself  --  would a real user genuinely be satisfied with ANY valid option, or do they most likely have a specific intent they failed to communicate?\ If you find yourself choosing one option among several equally plausible alternatives without a clear basis, that is a sign \texttt{u\_request} should be high.\\[2pt]
Then explain your assessment in \texttt{<u\_request\_}\allowbreak\texttt{explanation>}\,\ldots\,\texttt{</u\_request\_}\allowbreak\texttt{explanation>} tags.\\[2pt]
If \texttt{u\_request >= }$\theta$, your action MUST be \texttt{request\_clarification}.\\
Once you've finished your reasoning, you should choose an admissible action for the current step and present it within \texttt{<action> </action>} tags.
\end{quote}

\paragraph{History-entry template} Identical to Appendix~\ref{app:prompt-proposed-webshop}.

\paragraph{Confidence elicitation suffix} Identical to Appendix~\ref{app:prompt-react-webshop}.

\subsubsection{REAL}
\label{app:prompt-proposed-real}

\paragraph{System message}

\begin{quote}\small\itshape\raggedright
\# Instructions\\[2pt]
Review the current state of the page and all other information to find the best possible next action to accomplish your goal.\ Your answer will be interpreted and executed by a program, make sure to follow the formatting instructions.\\[2pt]
You should first reason step-by-step about the current situation.\ This reasoning process MUST be enclosed within \texttt{<think> </think>} tags.\\[2pt]
After thinking, assess request uncertainty in \texttt{<u\_request>}\,\ldots\,\texttt{</u\_request>} tags and explain in \texttt{<u\_request\_}\allowbreak\texttt{explanation>}\,\ldots\,\texttt{</u\_request\_}\allowbreak\texttt{explanation>} tags.\\[2pt]
Once you've finished your reasoning, you should choose an admissible action for the current step and present it within \texttt{<action> </action>} tags.\\[2pt]
After your action, provide confidence in \texttt{<confidence>}\,\ldots\,\texttt{</confidence>} tags and an explanation in \texttt{<explanation>}\,\ldots\,\texttt{</explanation>} tags.
\end{quote}

\paragraph{Next-action block}

\begin{quote}\small\itshape\raggedright
You are now at step \{t\}.\ Prior to this step, you have already taken \{t\} step(s).\ Now it's your turn to take an action.\\[2pt]
After thinking, you MUST assess your request uncertainty (0.0--1.0) in \texttt{<u\_request>...</u\_request>} tags.\\
\hspace*{1em}0.0 = the goal fully specifies every parameter  --  there is exactly one correct solution.\\
\hspace*{1em}0.5 = the goal leaves open at least one choice where the user likely has a specific preference they did not state  --  you would be guessing on their behalf.\\
\hspace*{1em}1.0 = critical details are missing, many equally valid interpretations exist.\\[2pt]
Be meticulous: if the goal leaves ANY parameter open-ended, ask yourself  --  would a real user genuinely be satisfied with ANY valid option, or do they most likely have a specific intent they failed to communicate?\ If you find yourself choosing one option among several equally plausible alternatives without a clear basis, that is a sign \texttt{u\_request} should be high.\\[2pt]
Then explain your assessment in \texttt{<u\_request\_}\allowbreak\texttt{explanation>}\,\ldots\,\texttt{</u\_request\_}\allowbreak\texttt{explanation>} tags.
\end{quote}

\paragraph{History-entry template} Identical to Appendix~\ref{app:prompt-proposed-webshop}.

\paragraph{Confidence elicitation suffix} Identical to Appendix~\ref{app:prompt-react-webshop}.

\section{Full Result Tables}
\label{app:tables}

Tables~\ref{tab:clarification_full}--\ref{tab:fault_clarification_full} report the complete per-backbone results summarized by the figures of Section~\ref{sec:results}.
Table~\ref{tab:clarification_full} lists the clarification-seeking metrics (success rate, precision, recall, F1, accuracy) for all five backbones on the two clarification-augmented benchmarks, expanding Figure~\ref{fig:clarification_bars}.
Table~\ref{tab:fault_standard_full} lists the fault-detection metrics (ROC-AUC, ECE, Brier) and success rate for every (backbone, method, aggregation) combination on the three standard benchmarks, expanding the GPT-5.1-only view of Figure~\ref{fig:fault_detection_bars} to all five backbones.
Table~\ref{tab:fault_clarification_full} reports the same fault-detection breakdown on the two clarification-augmented benchmarks.
For the proposed method, both fault-detection tables additionally include the request-uncertainty ($u_r$) aggregations of Section~\ref{sec:method-agg} alongside the action-confidence ($c$) aggregations.

\begin{table*}[t]
\centering
\caption{Clarification seeking: full metrics across backbones for both clarification-augmented benchmarks. Best per (backbone, benchmark, metric) in bold. The proposed method leads clarification F1 on every backbone on WebShop-Clar.\ and on four of five on ALFWorld-Clar., confirming that the decomposition enables clarification seeking where scalar-confidence baselines cannot.}
\label{tab:clarification_full}
\setlength{\tabcolsep}{3pt}
\footnotesize
\resizebox{\textwidth}{!}{%
\begin{tabular}{|l|l|ccccc|ccccc|}
\hline
 & & \multicolumn{5}{c|}{\textbf{WebShop-Clar.}} & \multicolumn{5}{c|}{\textbf{ALFWorld-Clar.}} \\
\textbf{Model} & \textbf{Method} & Success Rate (\%) & Precision & Recall & F1 & Accuracy & Success Rate (\%) & Precision & Recall & F1 & Accuracy \\
\hline
\multirow{3}{*}{GPT-5.1} & ReAct+UE & \textbf{24.5} & 0.235 & 0.085 & 0.125 & 0.404 & 36.2 & 0.643 & 0.257 & 0.367 & 0.551 \\
 & UAM & 18.9 & 0.238 & 0.102 & 0.143 & 0.368 & \textbf{47.1} & 0.737 & 0.424 & 0.538 & 0.647 \\
 & Proposed & 13.6 & \textbf{0.267} & \textbf{0.105} & \textbf{0.151} & \textbf{0.444} & 45.7 & \textbf{0.938} & \textbf{0.484} & \textbf{0.638} & \textbf{0.757} \\
\hline
\multirow{3}{*}{DeepSeek-v3.2-exp} & ReAct+UE & 7.0 & 0.139 & 0.100 & 0.116 & 0.240 & 54.0 & 0.675 & 0.900 & \textbf{0.771} & 0.746 \\
 & UAM & 4.0 & 0.173 & 0.180 & 0.176 & 0.160 & \textbf{56.0} & \textbf{0.848} & 0.700 & 0.767 & \textbf{0.798} \\
 & Proposed & \textbf{8.0} & \textbf{0.547} & \textbf{0.700} & \textbf{0.614} & \textbf{0.560} & 22.9 & 0.591 & \textbf{0.929} & 0.722 & 0.639 \\
\hline
\multirow{3}{*}{GLM-4.7} & ReAct+UE & 11.2 & 0.625 & 0.300 & 0.405 & 0.551 & 64.7 & \textbf{0.985} & 0.045 & 0.087 & 0.588 \\
 & UAM & 11.4 & 0.524 & 0.268 & 0.355 & 0.494 & 54.7 & 0.818 & 0.250 & 0.383 & 0.613 \\
 & Proposed & \textbf{19.0} & \textbf{0.694} & \textbf{0.829} & \textbf{0.756} & \textbf{0.722} & \textbf{68.9} & 0.737 & \textbf{0.609} & \textbf{0.667} & \textbf{0.689} \\
\hline
\multirow{3}{*}{Qwen3.5-35B} & ReAct+UE & 5.1 & 0.097 & 0.060 & 0.074 & 0.242 & \textbf{23.9} & \textbf{0.765} & 0.302 & 0.433 & 0.521 \\
 & UAM & 2.0 & 0.178 & 0.160 & 0.168 & 0.210 & 21.4 & 0.657 & 0.500 & 0.568 & 0.583 \\
 & Proposed & \textbf{6.4} & \textbf{0.500} & \textbf{0.511} & \textbf{0.505} & \textbf{0.500} & 20.6 & 0.561 & \textbf{0.742} & \textbf{0.639} & \textbf{0.618} \\
\hline
\multirow{3}{*}{GPT-OSS-120B} & ReAct+UE & 2.3 & 0.184 & 0.149 & 0.165 & 0.174 & 27.1 & 0.556 & 0.204 & 0.299 & 0.510 \\
 & UAM & 3.3 & 0.217 & 0.222 & 0.220 & \textbf{0.211} & 26.1 & 0.500 & 0.159 & 0.241 & 0.522 \\
 & Proposed & \textbf{4.4} & \textbf{0.245} & \textbf{0.255} & \textbf{0.250} & 0.209 & \textbf{28.4} & \textbf{0.714} & \textbf{0.729} & \textbf{0.722} & \textbf{0.716} \\
\hline
\end{tabular}}
\end{table*}

\begin{table*}[!htbp]
\centering
\caption{Full fault-detection results on the standard benchmarks across all backbones, methods, and trajectory-level aggregations. Success rate is method-level and shown once per (backbone, method). Best per (backbone, benchmark, metric) in bold. All three methods achieve comparable fault-detection ROC-AUC across backbones and aggregations, confirming that the request-uncertainty decomposition preserves the conventional uncertainty objective while additionally enabling clarification seeking.}
\label{tab:fault_standard_full}
\setlength{\tabcolsep}{2pt}
\tiny
\begin{tabular}{|l|l|l|cccc|cccc|cccc|}
\hline
 & & & \multicolumn{4}{c|}{\textbf{WebShop}} & \multicolumn{4}{c|}{\textbf{ALFWorld}} & \multicolumn{4}{c|}{\textbf{REAL}} \\
\textbf{Model} & \textbf{Method} & \textbf{Agg.} & ROC-AUC & ECE & Brier & Success Rate (\%) & ROC-AUC & ECE & Brier & Success Rate (\%) & ROC-AUC & ECE & Brier & Success Rate (\%) \\
\hline
\multirow{16}{*}{GPT-5.1} & \multirow{4}{*}{ReAct+UE} & c/last & 0.861 & 0.484 & 0.405 & 31.9 & 0.849 & 0.346 & 0.320 & 57.5 & 0.424 & 0.521 & 0.507 & \textbf{32.1} \\
 &  & c/avg & 0.783 & 0.433 & 0.373 &  & 0.645 & 0.300 & 0.319 &  & 0.526 & 0.500 & 0.470 &  \\
 &  & c/min & 0.775 & 0.236 & 0.232 &  & 0.712 & 0.134 & 0.235 &  & 0.564 & 0.397 & 0.374 &  \\
 &  & c/product & 0.781 & 0.171 & 0.203 &  & 0.909 & 0.378 & 0.300 &  & 0.503 & 0.357 & 0.347 &  \\
\cline{2-15}
 & \multirow{4}{*}{UAM} & c/last & 0.811 & 0.256 & 0.235 & \textbf{41.8} & 0.949 & 0.239 & 0.217 & \textbf{68.5} & 0.705 & 0.520 & 0.449 & 30.4 \\
 &  & c/avg & 0.856 & 0.284 & 0.255 &  & 0.759 & 0.179 & 0.224 &  & 0.732 & 0.502 & 0.445 &  \\
 &  & c/min & 0.838 & 0.151 & 0.172 &  & 0.749 & \textbf{0.074} & \textbf{0.186} &  & 0.661 & 0.392 & 0.351 &  \\
 &  & c/product & 0.849 & 0.317 & 0.275 &  & \textbf{0.962} & 0.441 & 0.327 &  & 0.455 & 0.226 & 0.296 &  \\
\cline{2-15}
 & \multirow{8}{*}{Proposed} & c/last & \textbf{0.902} & 0.447 & 0.348 & 26.1 & 0.940 & 0.339 & 0.312 & 58.9 & 0.733 & 0.610 & 0.512 & 19.2 \\
 &  & c/avg & 0.873 & 0.476 & 0.382 &  & 0.639 & 0.295 & 0.316 &  & \textbf{0.857} & 0.592 & 0.484 &  \\
 &  & c/min & 0.855 & 0.313 & 0.243 &  & 0.616 & 0.187 & 0.265 &  & 0.810 & 0.480 & 0.360 &  \\
 &  & c/product & 0.879 & \textbf{0.117} & \textbf{0.143} &  & 0.900 & 0.328 & 0.255 &  & 0.657 & \textbf{0.212} & \textbf{0.209} &  \\
 &  & $u_r$/first & 0.444 & 0.178 & 0.247 &  & 0.456 & 0.329 & 0.362 &  & 0.095 & 0.408 & 0.338 &  \\
 &  & $u_r$/max & 0.342 & 0.339 & 0.350 &  & 0.451 & 0.289 & 0.336 &  & 0.162 & 0.554 & 0.446 &  \\
 &  & $u_r$/avg & 0.344 & 0.280 & 0.298 &  & 0.620 & 0.400 & 0.387 &  & 0.124 & 0.514 & 0.366 &  \\
 &  & $u_r$/product & 0.725 & 0.256 & 0.257 &  & 0.841 & 0.589 & 0.588 &  & 0.152 & 0.309 & 0.259 &  \\
\hline
\multirow{16}{*}{DeepSeek-v3.2-exp} & \multirow{4}{*}{ReAct+UE} & c/last & 0.790 & 0.451 & 0.358 & \textbf{23.0} & 0.879 & 0.121 & \textbf{0.158} & \textbf{72.9} & \textbf{0.755} & 0.795 & 0.657 & 1.9 \\
 &  & c/avg & 0.779 & 0.478 & 0.385 &  & 0.849 & 0.140 & 0.174 &  & 0.725 & 0.799 & 0.658 &  \\
 &  & c/min & 0.732 & 0.218 & 0.207 &  & 0.809 & 0.166 & 0.164 &  & 0.353 & 0.703 & 0.528 &  \\
 &  & c/product & 0.853 & 0.184 & 0.181 &  & \textbf{0.983} & 0.620 & 0.548 &  & 0.216 & 0.395 & 0.248 &  \\
\cline{2-15}
 & \multirow{4}{*}{UAM} & c/last & 0.720 & 0.583 & 0.446 & 8.1 & 0.765 & 0.152 & 0.203 & 70.8 & 0.040 & 0.799 & 0.678 & 2.0 \\
 &  & c/avg & 0.758 & 0.600 & 0.432 &  & 0.667 & 0.091 & 0.205 &  & 0.120 & 0.801 & 0.673 &  \\
 &  & c/min & 0.731 & 0.350 & 0.196 &  & 0.584 & \textbf{0.071} & 0.211 &  & 0.100 & 0.744 & 0.593 &  \\
 &  & c/product & \textbf{0.890} & \textbf{0.068} & \textbf{0.054} &  & 0.899 & 0.610 & 0.543 &  & 0.020 & 0.638 & 0.463 &  \\
\cline{2-15}
 & \multirow{8}{*}{Proposed} & c/last & 0.698 & 0.444 & 0.347 & 15.1 & 0.689 & 0.127 & 0.201 & 70.9 & 0.591 & 0.665 & 0.569 & \textbf{13.0} \\
 &  & c/avg & 0.679 & 0.472 & 0.364 &  & 0.584 & 0.112 & 0.209 &  & 0.549 & 0.678 & 0.578 &  \\
 &  & c/min & 0.662 & 0.308 & 0.223 &  & 0.688 & 0.083 & 0.192 &  & 0.547 & 0.529 & 0.419 &  \\
 &  & c/product & 0.795 & 0.139 & 0.136 &  & 0.951 & 0.625 & 0.565 &  & 0.576 & 0.243 & 0.233 &  \\
 &  & $u_r$/first & 0.460 & 0.310 & 0.279 &  & 0.580 & 0.306 & 0.306 &  & 0.542 & 0.259 & 0.204 &  \\
 &  & $u_r$/max & 0.424 & 0.367 & 0.338 &  & 0.544 & 0.256 & 0.279 &  & 0.525 & 0.338 & 0.253 &  \\
 &  & $u_r$/avg & 0.387 & 0.340 & 0.319 &  & 0.533 & 0.270 & 0.292 &  & 0.514 & 0.287 & 0.221 &  \\
 &  & $u_r$/product & 0.604 & 0.142 & 0.135 &  & 0.836 & 0.694 & 0.681 &  & 0.505 & \textbf{0.174} & \textbf{0.146} &  \\
\hline
\multirow{16}{*}{GLM-4.7} & \multirow{4}{*}{ReAct+UE} & c/last & 0.860 & 0.459 & 0.343 & 23.0 & 0.762 & 0.123 & \textbf{0.131} & \textbf{83.1} & --- & --- & --- & 0.0 \\
 &  & c/avg & 0.944 & 0.482 & 0.341 &  & 0.666 & \textbf{0.055} & 0.138 &  & --- & --- & --- &  \\
 &  & c/min & 0.888 & 0.274 & 0.178 &  & 0.596 & 0.152 & 0.161 &  & --- & --- & --- &  \\
 &  & c/product & \textbf{0.975} & 0.185 & 0.167 &  & \textbf{0.862} & 0.554 & 0.437 &  & --- & --- & --- &  \\
\cline{2-15}
 & \multirow{4}{*}{UAM} & c/last & 0.799 & 0.448 & 0.336 & 17.8 & 0.698 & 0.213 & 0.225 & 70.9 & 0.690 & 0.621 & 0.549 & \textbf{22.4} \\
 &  & c/avg & 0.889 & 0.492 & 0.350 &  & 0.605 & 0.135 & 0.217 &  & 0.683 & 0.626 & 0.551 &  \\
 &  & c/min & 0.836 & 0.319 & 0.222 &  & 0.615 & 0.076 & 0.199 &  & 0.687 & 0.492 & 0.407 &  \\
 &  & c/product & 0.936 & \textbf{0.153} & \textbf{0.128} &  & 0.799 & 0.502 & 0.436 &  & 0.692 & 0.158 & 0.219 &  \\
\cline{2-15}
 & \multirow{8}{*}{Proposed} & c/last & 0.819 & 0.415 & 0.315 & \textbf{23.8} & 0.729 & 0.170 & 0.187 & 75.0 & \textbf{0.710} & 0.729 & 0.648 & 13.0 \\
 &  & c/avg & 0.812 & 0.422 & 0.332 &  & 0.633 & 0.119 & 0.186 &  & 0.659 & 0.721 & 0.634 &  \\
 &  & c/min & 0.730 & 0.283 & 0.243 &  & 0.604 & 0.078 & 0.183 &  & 0.697 & 0.583 & 0.457 &  \\
 &  & c/product & 0.869 & 0.189 & 0.197 &  & 0.792 & 0.461 & 0.387 &  & 0.650 & 0.234 & 0.200 &  \\
 &  & $u_r$/first & 0.397 & 0.231 & 0.308 &  & 0.526 & 0.398 & 0.374 &  & 0.340 & 0.236 & 0.186 &  \\
 &  & $u_r$/max & 0.339 & 0.451 & 0.448 &  & 0.480 & 0.367 & 0.358 &  & 0.373 & 0.326 & 0.277 &  \\
 &  & $u_r$/avg & 0.267 & 0.451 & 0.429 &  & 0.544 & 0.480 & 0.432 &  & 0.378 & 0.244 & 0.191 &  \\
 &  & $u_r$/product & 0.436 & 0.251 & 0.250 &  & 0.777 & 0.748 & 0.747 &  & 0.457 & \textbf{0.141} & \textbf{0.139} &  \\
\hline
\multirow{16}{*}{Qwen3.5-35B} & \multirow{4}{*}{ReAct+UE} & c/last & 0.866 & 0.523 & 0.407 & 15.0 & 0.899 & 0.432 & 0.370 & \textbf{45.1} & 0.605 & 0.796 & 0.732 & \textbf{9.0} \\
 &  & c/avg & 0.956 & 0.517 & 0.353 &  & 0.889 & 0.406 & 0.369 &  & 0.667 & 0.800 & 0.725 &  \\
 &  & c/min & 0.922 & 0.209 & 0.140 &  & 0.845 & 0.200 & 0.209 &  & 0.559 & 0.606 & 0.509 &  \\
 &  & c/product & 0.925 & \textbf{0.053} & 0.081 &  & \textbf{0.965} & 0.288 & 0.211 &  & 0.670 & 0.202 & 0.178 &  \\
\cline{2-15}
 & \multirow{4}{*}{UAM} & c/last & 0.925 & 0.562 & 0.428 & 18.0 & 0.536 & 0.547 & 0.534 & 38.9 & 0.528 & 0.860 & 0.806 & 5.0 \\
 &  & c/avg & 0.925 & 0.568 & 0.436 &  & 0.441 & 0.508 & 0.499 &  & 0.455 & 0.855 & 0.792 &  \\
 &  & c/min & 0.853 & 0.427 & 0.299 &  & 0.507 & 0.378 & 0.384 &  & 0.441 & 0.772 & 0.699 &  \\
 &  & c/product & 0.946 & 0.079 & 0.082 &  & 0.852 & 0.160 & \textbf{0.185} &  & 0.586 & 0.258 & 0.187 &  \\
\cline{2-15}
 & \multirow{8}{*}{Proposed} & c/last & 0.914 & 0.499 & 0.372 & \textbf{21.7} & 0.698 & 0.467 & 0.435 & 41.7 & \textbf{0.899} & 0.811 & 0.709 & 5.4 \\
 &  & c/avg & 0.887 & 0.488 & 0.369 &  & 0.761 & 0.400 & 0.377 &  & 0.652 & 0.802 & 0.698 &  \\
 &  & c/min & 0.853 & 0.371 & 0.267 &  & 0.715 & 0.292 & 0.302 &  & 0.452 & 0.727 & 0.601 &  \\
 &  & c/product & \textbf{0.959} & 0.073 & \textbf{0.073} &  & 0.921 & 0.256 & 0.208 &  & 0.880 & 0.154 & 0.114 &  \\
 &  & $u_r$/first & 0.372 & 0.272 & 0.322 &  & 0.493 & \textbf{0.104} & 0.260 &  & 0.792 & 0.377 & 0.231 &  \\
 &  & $u_r$/max & 0.195 & 0.519 & 0.517 &  & 0.511 & 0.110 & 0.265 &  & 0.756 & 0.401 & 0.259 &  \\
 &  & $u_r$/avg & 0.288 & 0.440 & 0.436 &  & 0.413 & 0.221 & 0.283 &  & 0.398 & 0.296 & 0.203 &  \\
 &  & $u_r$/product & 0.521 & 0.222 & 0.227 &  & 0.814 & 0.413 & 0.410 &  & 0.487 & \textbf{0.064} & \textbf{0.065} &  \\
\hline
\multirow{16}{*}{GPT-OSS-120B} & \multirow{4}{*}{ReAct+UE} & c/last & 0.862 & 0.696 & 0.559 & \textbf{8.9} & 0.860 & 0.478 & 0.441 & \textbf{44.4} & 0.718 & 0.728 & 0.620 & 10.0 \\
 &  & c/avg & 0.892 & 0.711 & 0.579 &  & 0.709 & 0.456 & 0.443 &  & 0.622 & 0.711 & 0.596 &  \\
 &  & c/min & 0.844 & 0.539 & 0.364 &  & 0.638 & 0.301 & 0.323 &  & 0.566 & 0.536 & 0.393 &  \\
 &  & c/product & 0.950 & 0.069 & 0.068 &  & \textbf{0.888} & 0.234 & 0.209 &  & 0.677 & 0.179 & 0.162 &  \\
\cline{2-15}
 & \multirow{4}{*}{UAM} & c/last & 0.869 & 0.769 & 0.653 & 7.8 & 0.715 & 0.627 & 0.606 & 33.3 & 0.522 & 0.713 & 0.645 & \textbf{15.0} \\
 &  & c/avg & 0.917 & 0.773 & 0.662 &  & 0.524 & 0.590 & 0.568 &  & 0.568 & 0.707 & 0.627 &  \\
 &  & c/min & 0.718 & 0.677 & 0.519 &  & 0.568 & 0.471 & 0.442 &  & 0.687 & 0.559 & 0.435 &  \\
 &  & c/product & \textbf{0.985} & \textbf{0.042} & \textbf{0.028} &  & 0.763 & 0.125 & 0.184 &  & \textbf{0.749} & \textbf{0.128} & \textbf{0.146} &  \\
\cline{2-15}
 & \multirow{8}{*}{Proposed} & c/last & 0.826 & 0.825 & 0.742 & 6.9 & 0.591 & 0.629 & 0.612 & 33.7 & 0.636 & 0.774 & 0.711 & 13.0 \\
 &  & c/avg & 0.747 & 0.814 & 0.725 &  & 0.453 & 0.595 & 0.579 &  & 0.711 & 0.763 & 0.691 &  \\
 &  & c/min & 0.765 & 0.704 & 0.557 &  & 0.592 & 0.497 & 0.463 &  & 0.687 & 0.656 & 0.541 &  \\
 &  & c/product & 0.940 & 0.067 & 0.041 &  & 0.826 & \textbf{0.094} & \textbf{0.158} &  & 0.666 & 0.312 & 0.260 &  \\
 &  & $u_r$/first & 0.227 & 0.293 & 0.210 &  & 0.452 & 0.254 & 0.304 &  & 0.270 & 0.366 & 0.317 &  \\
 &  & $u_r$/max & 0.258 & 0.485 & 0.397 &  & 0.403 & 0.352 & 0.376 &  & 0.204 & 0.559 & 0.447 &  \\
 &  & $u_r$/avg & 0.382 & 0.248 & 0.206 &  & 0.556 & 0.194 & 0.263 &  & 0.267 & 0.395 & 0.281 &  \\
 &  & $u_r$/product & 0.660 & 0.070 & 0.069 &  & 0.739 & 0.332 & 0.328 &  & 0.421 & 0.211 & 0.202 &  \\
\hline
\end{tabular}
\end{table*}

\begin{table*}[!htbp]
\centering
\caption{Full fault-detection results on the clarification-augmented benchmarks across all backbones, methods, and aggregations. For the proposed method we additionally report the $u_r$-based score variants. Success rate is method-level and shown once per (backbone, method). Best per (backbone, benchmark, metric) in bold. $u_r$-based aggregations are systematically weaker on fault-detection metrics than $c$-based aggregations, confirming that request uncertainty is designed for clarification seeking rather than predicting trajectory failure.}
\label{tab:fault_clarification_full}
\setlength{\tabcolsep}{2pt}
\tiny
\begin{tabular}{|l|l|l|cccc|cccc|}
\hline
 & & & \multicolumn{4}{c|}{\textbf{WebShop-Clar.}} & \multicolumn{4}{c|}{\textbf{ALFWorld-Clar.}} \\
\textbf{Model} & \textbf{Method} & \textbf{Agg.} & ROC-AUC & ECE & Brier & Success Rate (\%) & ROC-AUC & ECE & Brier & Success Rate (\%) \\
\hline
\multirow{16}{*}{GPT-5.1} & \multirow{4}{*}{ReAct+UE} & c/last & 0.628 & 0.642 & 0.588 & \textbf{24.5} & 0.865 & 0.537 & 0.496 & 36.2 \\
 &  & c/avg & 0.370 & 0.576 & 0.530 &  & 0.548 & 0.507 & 0.485 &  \\
 &  & c/min & 0.295 & 0.459 & 0.451 &  & 0.581 & 0.352 & 0.346 &  \\
 &  & c/product & 0.312 & 0.436 & 0.365 &  & 0.916 & \textbf{0.223} & \textbf{0.199} &  \\
\cline{2-11}
 & \multirow{4}{*}{UAM} & c/last & 0.461 & 0.658 & 0.611 & 18.9 & 0.814 & 0.437 & 0.411 & \textbf{47.1} \\
 &  & c/avg & 0.356 & 0.613 & 0.550 &  & 0.590 & 0.392 & 0.402 &  \\
 &  & c/min & 0.356 & 0.479 & 0.449 &  & 0.456 & 0.264 & 0.330 &  \\
 &  & c/product & 0.336 & 0.421 & 0.364 &  & 0.910 & 0.336 & 0.286 &  \\
\cline{2-11}
 & \multirow{8}{*}{Proposed} & c/last & \textbf{0.890} & 0.656 & 0.537 & 13.6 & 0.862 & 0.453 & 0.418 & 45.7 \\
 &  & c/avg & 0.869 & 0.663 & 0.543 &  & 0.708 & 0.420 & 0.413 &  \\
 &  & c/min & 0.746 & 0.561 & 0.427 &  & 0.630 & 0.321 & 0.336 &  \\
 &  & c/product & 0.634 & 0.199 & 0.181 &  & \textbf{0.947} & 0.273 & 0.219 &  \\
 &  & $u_r$/first & 0.319 & 0.133 & 0.152 &  & 0.366 & 0.311 & 0.389 &  \\
 &  & $u_r$/max & 0.205 & 0.264 & 0.252 &  & 0.352 & 0.345 & 0.400 &  \\
 &  & $u_r$/avg & 0.127 & 0.275 & 0.187 &  & 0.445 & 0.348 & 0.372 &  \\
 &  & $u_r$/product & 0.497 & \textbf{0.126} & \textbf{0.136} &  & 0.694 & 0.457 & 0.457 &  \\
\hline
\multirow{16}{*}{DeepSeek-v3.2-exp} & \multirow{4}{*}{ReAct+UE} & c/last & 0.563 & 0.713 & 0.584 & 7.0 & 0.810 & 0.277 & 0.279 & 54.0 \\
 &  & c/avg & 0.462 & 0.720 & 0.589 &  & 0.593 & 0.235 & 0.296 &  \\
 &  & c/min & 0.473 & 0.548 & 0.403 &  & 0.699 & 0.103 & \textbf{0.232} &  \\
 &  & c/product & 0.485 & 0.292 & 0.201 &  & 0.825 & 0.475 & 0.435 &  \\
\cline{2-11}
 & \multirow{4}{*}{UAM} & c/last & 0.302 & 0.841 & 0.762 & 4.0 & 0.610 & 0.309 & 0.329 & \textbf{56.0} \\
 &  & c/avg & 0.372 & 0.768 & 0.636 &  & 0.518 & 0.234 & 0.302 &  \\
 &  & c/min & 0.469 & 0.530 & 0.356 &  & 0.581 & \textbf{0.062} & 0.242 &  \\
 &  & c/product & 0.497 & 0.229 & 0.153 &  & \textbf{0.842} & 0.501 & 0.465 &  \\
\cline{2-11}
 & \multirow{8}{*}{Proposed} & c/last & 0.577 & 0.706 & 0.609 & \textbf{8.0} & 0.725 & 0.491 & 0.428 & 22.9 \\
 &  & c/avg & 0.503 & 0.723 & 0.605 &  & 0.491 & 0.540 & 0.486 &  \\
 &  & c/min & 0.567 & 0.506 & 0.369 &  & 0.523 & 0.401 & 0.384 &  \\
 &  & c/product & \textbf{0.685} & 0.097 & 0.100 &  & 0.679 & 0.230 & 0.244 &  \\
 &  & $u_r$/first & 0.501 & 0.255 & 0.203 &  & 0.440 & 0.507 & 0.417 &  \\
 &  & $u_r$/max & 0.445 & 0.380 & 0.282 &  & 0.394 & 0.554 & 0.469 &  \\
 &  & $u_r$/avg & 0.356 & 0.309 & 0.233 &  & 0.139 & 0.542 & 0.458 &  \\
 &  & $u_r$/product & 0.444 & \textbf{0.079} & \textbf{0.081} &  & 0.215 & 0.236 & 0.241 &  \\
\hline
\multirow{16}{*}{GLM-4.7} & \multirow{4}{*}{ReAct+UE} & c/last & 0.690 & 0.666 & 0.562 & 11.2 & 0.629 & 0.286 & 0.298 & 64.7 \\
 &  & c/avg & 0.560 & 0.653 & 0.541 &  & 0.431 & 0.229 & 0.292 &  \\
 &  & c/min & 0.504 & 0.472 & 0.387 &  & 0.561 & \textbf{0.104} & \textbf{0.217} &  \\
 &  & c/product & 0.569 & 0.284 & 0.252 &  & 0.711 & 0.406 & 0.360 &  \\
\cline{2-11}
 & \multirow{4}{*}{UAM} & c/last & 0.646 & 0.596 & 0.485 & 11.4 & 0.674 & 0.345 & 0.343 & 54.7 \\
 &  & c/avg & 0.477 & 0.634 & 0.526 &  & 0.506 & 0.304 & 0.344 &  \\
 &  & c/min & 0.375 & 0.508 & 0.410 &  & 0.573 & 0.172 & 0.271 &  \\
 &  & c/product & 0.449 & 0.319 & 0.266 &  & 0.783 & 0.395 & 0.348 &  \\
\cline{2-11}
 & \multirow{8}{*}{Proposed} & c/last & \textbf{0.714} & 0.534 & 0.436 & \textbf{19.0} & 0.515 & 0.282 & 0.281 & \textbf{68.9} \\
 &  & c/avg & 0.682 & 0.568 & 0.468 &  & 0.403 & 0.218 & 0.267 &  \\
 &  & c/min & 0.559 & 0.418 & 0.343 &  & 0.470 & 0.131 & 0.222 &  \\
 &  & c/product & 0.677 & \textbf{0.208} & \textbf{0.201} &  & 0.662 & 0.411 & 0.365 &  \\
 &  & $u_r$/first & 0.546 & 0.209 & 0.238 &  & 0.402 & 0.404 & 0.445 &  \\
 &  & $u_r$/max & 0.533 & 0.430 & 0.383 &  & 0.433 & 0.387 & 0.418 &  \\
 &  & $u_r$/avg & 0.483 & 0.229 & 0.270 &  & 0.495 & 0.551 & 0.522 &  \\
 &  & $u_r$/product & 0.466 & 0.215 & 0.211 &  & \textbf{0.873} & 0.689 & 0.689 &  \\
\hline
\multirow{16}{*}{Qwen3.5-35B} & \multirow{4}{*}{ReAct+UE} & c/last & 0.615 & 0.836 & 0.759 & 5.1 & 0.844 & 0.628 & 0.548 & \textbf{23.9} \\
 &  & c/avg & 0.473 & 0.818 & 0.724 &  & 0.740 & 0.605 & 0.534 &  \\
 &  & c/min & 0.448 & 0.705 & 0.587 &  & 0.778 & 0.392 & 0.298 &  \\
 &  & c/product & 0.441 & 0.418 & 0.325 &  & \textbf{0.960} & 0.142 & \textbf{0.123} &  \\
\cline{2-11}
 & \multirow{4}{*}{UAM} & c/last & 0.235 & 0.932 & 0.894 & 2.0 & 0.604 & 0.720 & 0.685 & 21.4 \\
 &  & c/avg & 0.342 & 0.889 & 0.814 &  & 0.390 & 0.671 & 0.624 &  \\
 &  & c/min & 0.403 & 0.771 & 0.631 &  & 0.480 & 0.495 & 0.441 &  \\
 &  & c/product & 0.477 & 0.473 & 0.344 &  & 0.783 & \textbf{0.102} & 0.147 &  \\
\cline{2-11}
 & \multirow{8}{*}{Proposed} & c/last & \textbf{0.756} & 0.835 & 0.756 & \textbf{6.4} & 0.694 & 0.650 & 0.578 & 20.6 \\
 &  & c/avg & 0.570 & 0.821 & 0.737 &  & 0.530 & 0.620 & 0.549 &  \\
 &  & c/min & 0.516 & 0.701 & 0.566 &  & 0.479 & 0.535 & 0.459 &  \\
 &  & c/product & 0.754 & 0.269 & 0.185 &  & 0.835 & 0.120 & 0.126 &  \\
 &  & $u_r$/first & 0.237 & 0.245 & 0.156 &  & 0.372 & 0.346 & 0.276 &  \\
 &  & $u_r$/max & 0.144 & 0.571 & 0.493 &  & 0.443 & 0.346 & 0.293 &  \\
 &  & $u_r$/avg & 0.155 & 0.473 & 0.385 &  & 0.210 & 0.397 & 0.283 &  \\
 &  & $u_r$/product & 0.322 & \textbf{0.121} & \textbf{0.125} &  & 0.407 & 0.205 & 0.205 &  \\
\hline
\multirow{16}{*}{GPT-OSS-120B} & \multirow{4}{*}{ReAct+UE} & c/last & 0.482 & 0.861 & 0.771 & 2.3 & 0.821 & 0.637 & 0.584 & 27.1 \\
 &  & c/avg & 0.458 & 0.848 & 0.747 &  & 0.516 & 0.632 & 0.598 &  \\
 &  & c/min & 0.521 & 0.731 & 0.579 &  & 0.535 & 0.472 & 0.421 &  \\
 &  & c/product & 0.512 & 0.395 & 0.322 &  & \textbf{0.843} & \textbf{0.110} & \textbf{0.152} &  \\
\cline{2-11}
 & \multirow{4}{*}{UAM} & c/last & 0.420 & 0.905 & 0.855 & 3.3 & 0.610 & 0.693 & 0.672 & 26.1 \\
 &  & c/avg & 0.437 & 0.885 & 0.818 &  & 0.355 & 0.677 & 0.662 &  \\
 &  & c/min & 0.431 & 0.789 & 0.666 &  & 0.432 & 0.576 & 0.540 &  \\
 &  & c/product & 0.525 & 0.366 & 0.305 &  & 0.559 & 0.226 & 0.245 &  \\
\cline{2-11}
 & \multirow{8}{*}{Proposed} & c/last & 0.585 & 0.917 & 0.883 & \textbf{4.4} & 0.598 & 0.695 & 0.680 & \textbf{28.4} \\
 &  & c/avg & 0.536 & 0.907 & 0.865 &  & 0.396 & 0.673 & 0.658 &  \\
 &  & c/min & 0.520 & 0.854 & 0.773 &  & 0.518 & 0.605 & 0.568 &  \\
 &  & c/product & \textbf{0.661} & 0.478 & 0.339 &  & 0.729 & 0.185 & 0.195 &  \\
 &  & $u_r$/first & 0.509 & 0.146 & 0.129 &  & 0.406 & 0.428 & 0.432 &  \\
 &  & $u_r$/max & 0.273 & 0.532 & 0.477 &  & 0.411 & 0.437 & 0.458 &  \\
 &  & $u_r$/avg & 0.256 & 0.313 & 0.224 &  & 0.484 & 0.304 & 0.308 &  \\
 &  & $u_r$/product & 0.430 & \textbf{0.049} & \textbf{0.046} &  & 0.738 & 0.291 & 0.286 &  \\
\hline
\end{tabular}
\end{table*}

\section{Calibration Plots}
\label{app:calibration}

Reliability diagrams for the three methods (rows) across the five benchmarks (columns) under last-step aggregation, one figure per backbone.
Each point bins trajectories by action confidence and plots the observed success rate against the bin's predicted confidence; points below the diagonal indicate overconfidence.
The GPT-5.1 diagram appears in Section~\ref{sec:res-calib}; the remaining four backbones are shown here.

\begin{figure*}[!t]
\centering
\includegraphics[width=0.92\textwidth]{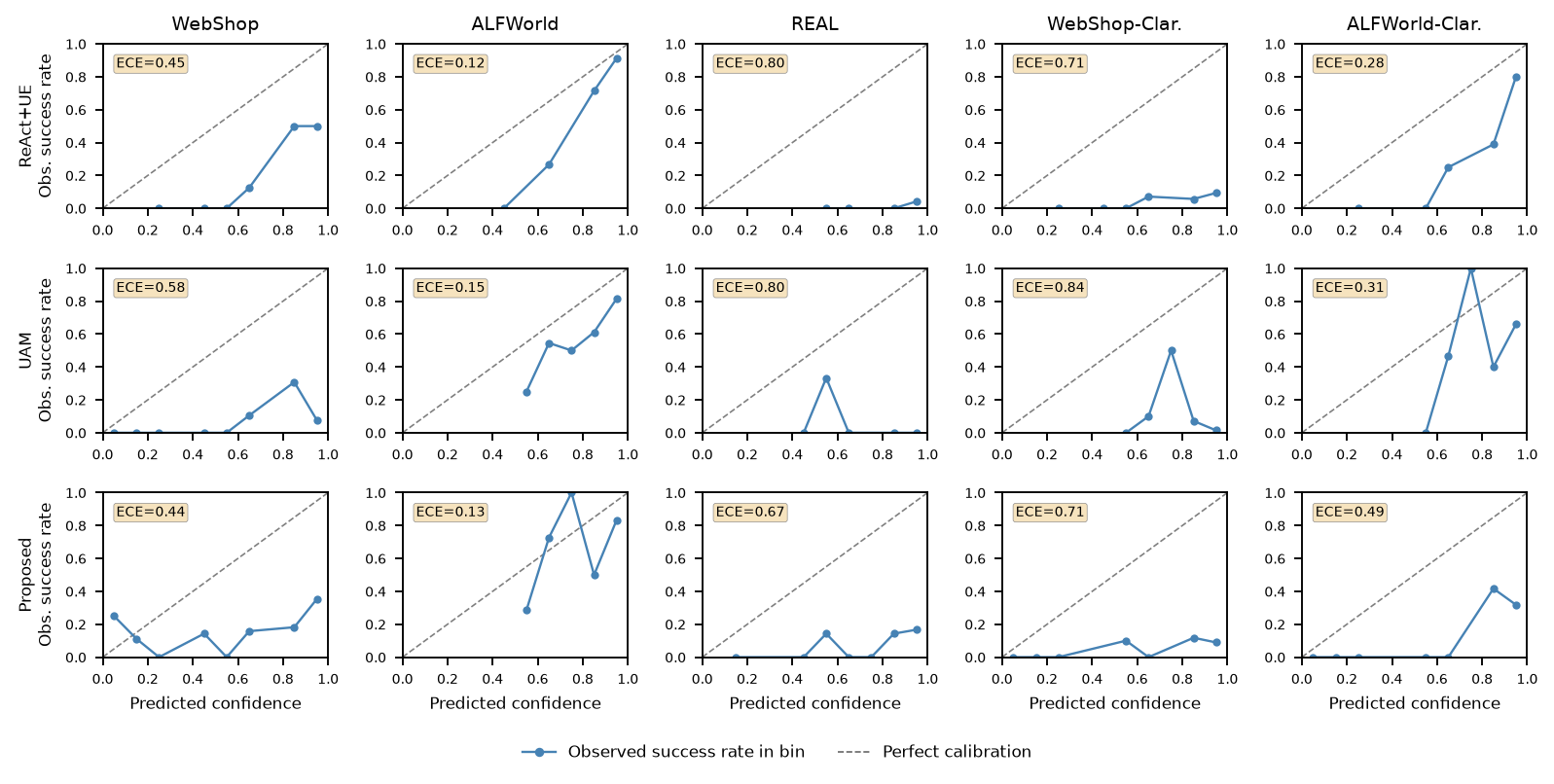}
\caption{Reliability diagrams for DeepSeek-v3.2-exp: the three methods (rows) across the five benchmarks (columns), under last-step aggregation.
All curves lie below the diagonal, indicating systematic overconfidence for every method and benchmark.}
\label{fig:calibration-deepseek}
\end{figure*}

\begin{figure*}[!t]
\centering
\includegraphics[width=0.92\textwidth]{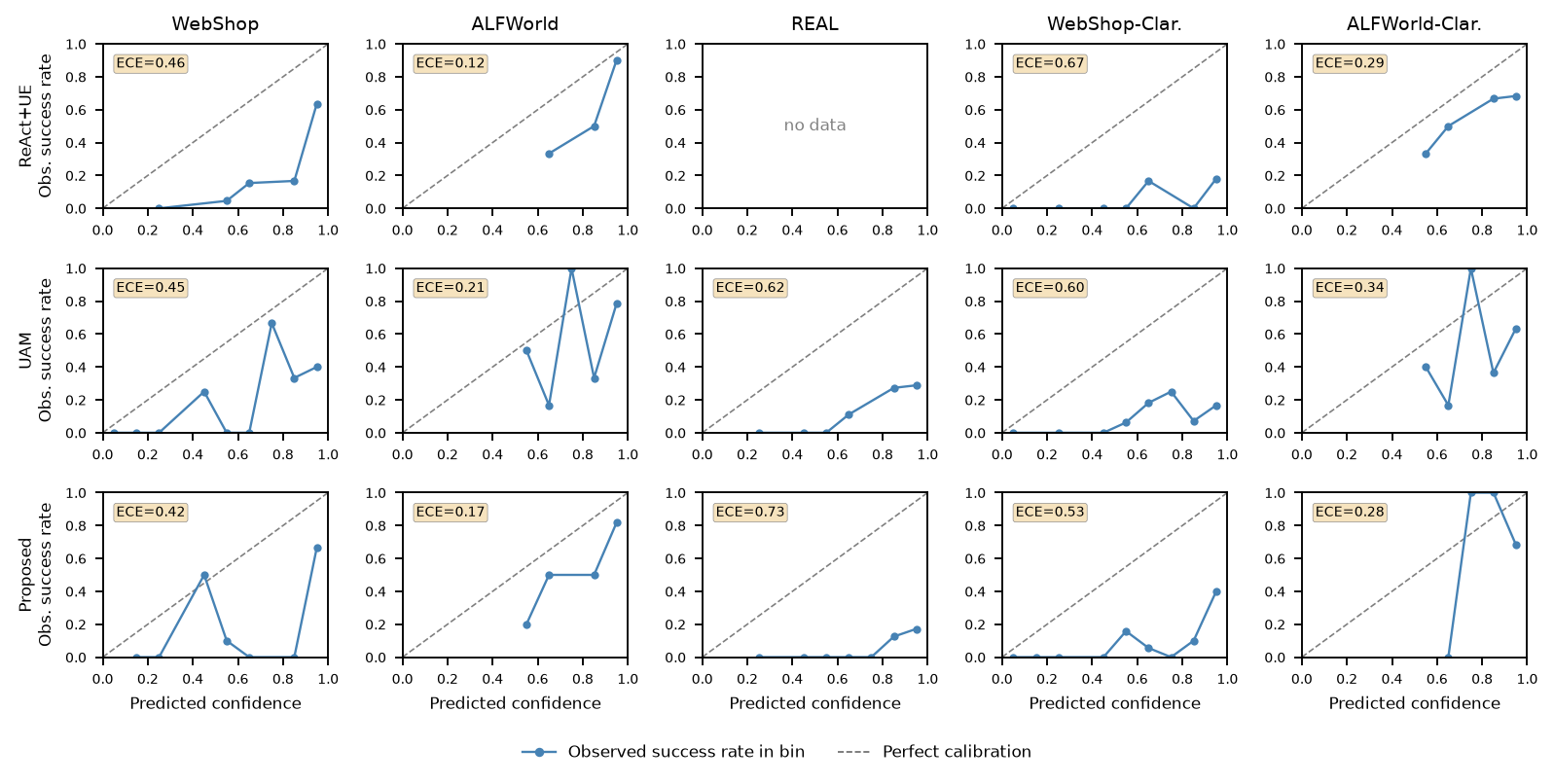}
\caption{Reliability diagrams for GLM-4.7: the three methods (rows) across the five benchmarks (columns), under last-step aggregation.
All curves lie below the diagonal, indicating systematic overconfidence for every method and benchmark.}
\label{fig:calibration-glm}
\end{figure*}

\begin{figure*}[!t]
\centering
\includegraphics[width=0.92\textwidth]{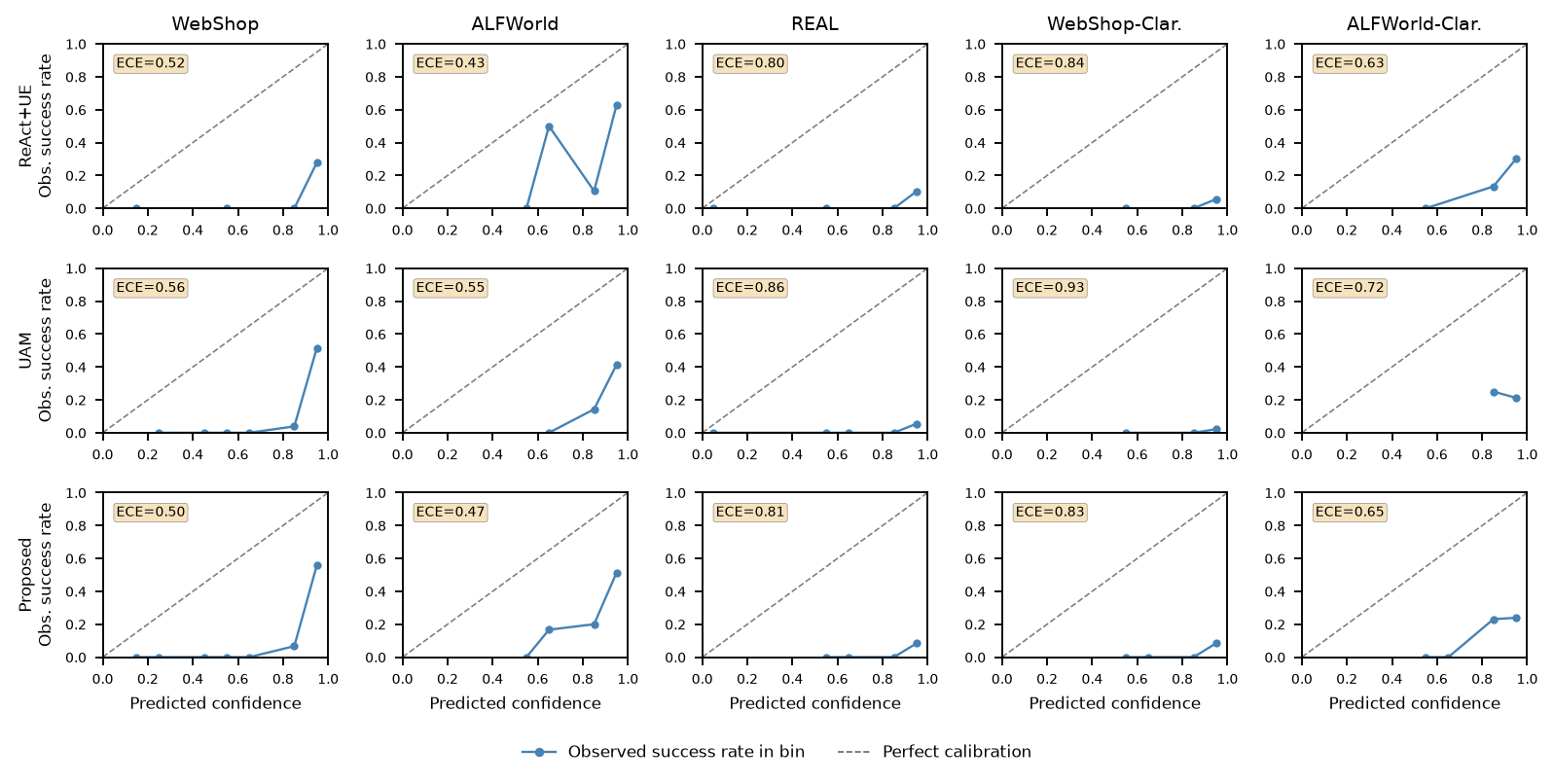}
\caption{Reliability diagrams for Qwen3.5-35B: the three methods (rows) across the five benchmarks (columns), under last-step aggregation.
All curves lie below the diagonal, indicating systematic overconfidence for every method and benchmark.}
\label{fig:calibration-qwen}
\end{figure*}

\begin{figure*}[!t]
\centering
\includegraphics[width=0.92\textwidth]{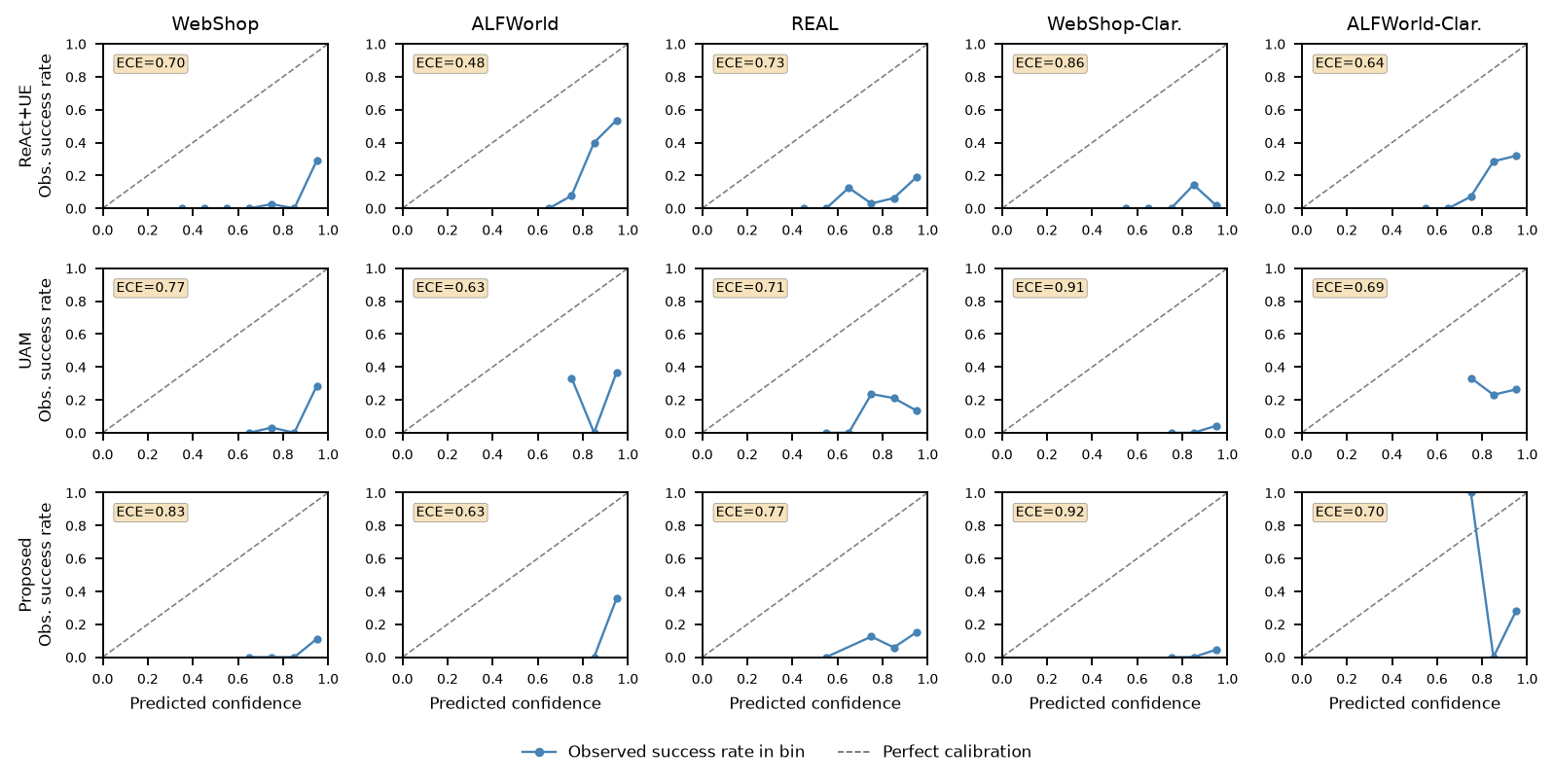}
\caption{Reliability diagrams for GPT-OSS-120B: the three methods (rows) across the five benchmarks (columns), under last-step aggregation.
All curves lie below the diagonal, indicating systematic overconfidence for every method and benchmark.}
\label{fig:calibration-oss}
\end{figure*}

\FloatBarrier
\bibliographystyle{IEEEtran}
\bibliography{references}

\end{document}